\documentclass{article}

\usepackage{style}

\usepackage{natbib} % has a nice set of citation styles and commands
\usepackage{mathtools} % amsmath with fixes and additions
\usepackage{booktabs} % commands to create good-looking tables
\usepackage{tikz} % nice language for creating drawings and diagrams

\usepackage{amssymb}
\usepackage{amsthm}
\usepackage{amsfonts}

\usepackage{dsfont}
\usepackage{subfiles}
\usepackage{bbm}
\usepackage{comment}
\usepackage{caption}
\usepackage{hyperref}
\usepackage{thm-restate}
\usepackage{wrapfig}

\usepackage{microtype}
\usepackage{graphicx}
\usepackage{subcaption}
\usepackage{booktabs} % for professional tables
\usepackage{arydshln}
\usepackage{algorithm}
\usepackage{algorithmic}
\usepackage{forloop}

\usepackage{blindtext}
\usepackage{multicol}

\usepackage[utf8]{inputenc} % allow utf-8 input
\usepackage[T1]{fontenc}    % use 8-bit T1 fonts
\usepackage{hyperref}       % hyperlinks
\usepackage{url}            % simple URL typesetting
\usepackage{booktabs}       % professional-quality tables
\usepackage{amsfonts}       % blackboard math symbols
\usepackage{nicefrac}       % compact symbols for 1/2, etc.
\usepackage{microtype}      % microtypography
\usepackage{xcolor}         % colors

\newtheorem{theorem}{Theorem}%[section]

\newtheorem{corollary}{Corollary}%[theorem]
\newtheorem{lemma}{Lemma}

\newtheorem{proposition}{Proposition}
\newenvironment{customprp}[1]{\renewcommand\theproposition{#1}\proposition}{\endproposition}
\newtheorem{definition}{Definition}
\newtheorem{remark}{Remark}

\newtheorem{claim}{Claim}
\newtheorem*{lemma*}{Lemma}

\usetikzlibrary{positioning}

\newcommand{\Pa}[2]{\textit{Pa}_{#2}(#1)}
\newcommand{\Ch}[2]{\textit{Ch}_{#2}(#1)}
\newcommand{\Anc}[2]{\textit{Anc}_{#2}(#1)}
\newcommand{\De}[2]{\textit{De}_{#2}(#1)}

\newcommand{\V}[0]{\mathbf{V}}
\newcommand{\W}[0]{\mathbf{W}}
\newcommand{\U}[0]{\mathbf{U}}

\newcommand{\X}[0]{\mathbf{X}}

\newcommand{\Y}[0]{\mathbf{Y}}
\newcommand{\Z}[0]{\mathbf{Z}}

\newcommand{\A}[0]{\mathbf{A}}
\newcommand{\B}[0]{\mathbf{B}}
\newcommand{\T}[0]{\mathbf{T}}

\newcommand{\N}[0]{\mathbf{N}}

\renewcommand{\H}{\mathbf{H}}
\renewcommand{\S}{\mathbf{S}}

\newcommand{\x}[0]{\mathbf{x}}
\newcommand{\y}[0]{\mathbf{y}}
\newcommand{\z}[0]{\mathbf{z}}
\newcommand{\w}[0]{\mathbf{w}}
\newcommand{\n}[0]{\mathbf{n}}
\newcommand{\e}[0]{\mathbf{e}}

\renewcommand{\a}[0]{\mathbf{a}}
\renewcommand{\b}[0]{\mathbf{b}}
\renewcommand{\d}[0]{\mathbf{d}}

\newcommand{\C}[0]{\mathcal{C}}
\newcommand{\G}[0]{\mathcal{G}}
\newcommand{\M}[0]{\mathcal{M}}
\newcommand{\F}[0]{\mathcal{F}}

\newcommand{\dom}[2]{\mathfrak{X}_{#2}(#1)}

\newcommand{\independent}{\perp\!\!\!\perp}

\title{On the Incompleteness of do-calculus for Universal Identifiability}

\date{} 					% Or removing it

\usepackage{authblk}

\author[1]{\href{mailto:26yaroslav@gmail.com}{Yaroslav Kivva}}
\author[2,1]{\href{mailto:a.novik24@imperial.ac.uk}{Aliaksandra Novik}}
\author[1]{Negar Kiyavash}
\affil[1]{%
    EPFL\\
    Lausanne, Switzerland
}
\affil[2]{%
    Imperial College\\
    London, United Kingdom
}

\begin{document}

\maketitle

\begin{abstract}
    We study the problem of causal effect identification from an \textit{arbitrary set of observational and/or interventional distributions}, which we call \textit{Universal Identifiability} (uID). The uID problem generalizes existing classical identifiability (ID) and general identifiability (gID) problems, which considered the identification problem from either mere observational distributions or a specific type of interventional distributions. Prior research has established graphical criteria to determine when a causal effect is identifiable in ID and gID problems. Moreover, algorithms that leverage do-calculus and basic probability manipulations have been proposed to compute these effects. Consequently, it has been shown that do-calculus is both \textit{sound} and \textit{complete} in these problems. Despite these well-established results, a formal definition of what it means for the do-calculus to be complete does not exist. We propose a rigorous formalization for the completeness of the do-calculus and show that the completeness results in previous work remain valid. However, for more general problems, such as uID, the rules of do-calculus are \textbf{no longer complete}. This is perhaps the reason why the uID problem has remained open thus far.
\end{abstract}

\section{Introduction}\label{sec:intro}
% Importance of causal effect identifiability
% About identifiability and do-calculus rules
% Previous research on Identifiability problems
% Our contribution
Estimating the causal effect of one set of random variables on another is a fundamental challenge in various fields, including artificial intelligence, social sciences, healthcare, and economics. To solve this problem, researchers collect observational and experimental data and harness the independencies among the variables of interest (often encoded in a directed acyclic graph (DAG) \citep{pearl1995causal}). When all the variables in the causal DAG are observed, then the causal effect $P(\y |do(\x))$ can be calculated uniquely from the observational distribution \citep{geiger1990logic}. However, if some variables are not observed, then the causal effect $P(\y|do(\x))$ may not be identifiable (i.e., it cannot be computed uniquely). Thus, understanding the conditions under which $P(\y |do(\x))$ is identifiable is widely studied in the literature.

In his seminal work, \cite{pearl1995causal} introduced $do$-calculus rules that allow one to manipulate probability functions $P(\cdot)$ containing interventions $do(\cdot)$. To this day, $do$-calculus rules are the backbone of all causal effect identification algorithms in DAGs. Sufficient conditions for identifiability were proposed in \citep{pearl1995causal, pearl1995probabilistic, kuroki1999identifiability, spirtes2001causation, qlemmas, tian2003ID, tian2004identifying}. These works later culminated in the well-known classical identifiability (ID) result of \cite{shpitser2006identification} and \cite{huang2006identifiability}. Specifically, they proposed a necessary and sufficient criterion to determine whether $P(\y|do(\x))$ is identifiable in a given causal DAG $\G$ from an observational distribution $P(\N)$. Furthermore, they proposed a procedure to compute the causal effect of interest from $P(\N)$ and, based on it, concluded that the rules of the $do$-calculus are sound and complete for this problem. In addition, \cite{shpitser2012identification, bareinboim2012causal, lee2019general, lee2020general, correa2021nested, kivva2022revisiting, kivva2023identifiability} extend the results of the ID problem by targeting a conditional causal effect $P(\y|do(\x), \z)$ and/or incorporating multiple observational distributions with a specific structure. Similarly, they provided necessary and sufficient conditions under which the target causal effect is identifiable, along with an algorithm to compute it. As a result, they concluded that $do$-calculus rules remain sound and complete for the problem they studied. The identifiability problem has been studied in the presence of side information of different types. For example, \cite{tikka2019causal} considered the case where the observations consist of multiple incomplete data sources, and \cite{abouei2024s, abouei2024causal} studied the problem under selection bias.

The issue with the treatment of completeness in the previous work is that the completeness of $do$-calculus is proved by proposing a \textit{specific} sound and complete algorithm. This algorithm can compute the causal effect as long as it is identifiable by a sequence of applications of $do$-calculus rules and standard probabilistic manipulations. That is, if the effect is identifiable, the algorithm returns a valid formula to compute, and if no formula is returned, then the effect must have been unidentifiable. Therefore, the limitation of this approach is that it can only be used to prove completeness of $do$-calculus rules through first identifying a \textit{specific} sound and complete algorithm, and in the absence of such an algorithm, we cannot establish whether the rules of $do$-calculus are complete or not.

In this work, we address the conjecture that the $do$-calculus is complete in the sense that if a causal effect is identifiable, then it can be computed from a sequence of applying the $do$-calculus rules. We do so by providing a formal definition of completeness of $do$-calculus for any arbitrary identifiability problem. Remarkably, to the best of our knowledge, no explicit formal definition of completeness for a general setting exists in the literature. As we shall see, our definition is consistent with the notions of completeness in the previously studied problems such as classical identifiability (ID) and general identifiability (gID). We consider an extension of the general identifiability (gID) where the restrictions on the graphical structure of observed distributions are relaxed. We show that in this extension problem, called universal identifiability (uID), the rules of $do$-calculus are \textbf{no longer complete}.

%To prove this statement, first,   %Further, As far we aware of, this is the \textbf{first} such result in the literature. More importantly, it illustrates the limits and complexity of problems that go beyond the gID setting.

\section{Preliminaries}
\subsection{Notation}
Throughout the paper, we use capital letters (e.g. $X$) to denote random variables and their corresponding lowercase letters (e.g. $x$) for realizations. Similarly, we use bold capital letters (e.g. $\X$) to represent a set of random variables (and their multivariate distribution) and bold lowercase letters (e.g. $\x$) for a specific realization of their multivariate distribution. Additionally, we denote by $\x_{\S}$ the entries in $\x$ that correspond to the variables in $\S$, i.e. the realization of the variables $\X\cap \S$.
Furthermore, we use $\dom{X}{}$ for the domain of random variable $X$ and consequently $\dom{\X}{}$ for the domain of the random vector $\X$ which is equal to the Cartesian product $\prod_{X \in \X}\dom{X}{}$.

This study assumes that the available data is generated by some probabilistic structural causal model (SCM) $\M$. The structural causal model $\M = \langle \G, \{f_{V}(\cdot)\}_{V\in \V}, \{\epsilon_{V}\}_{V\in \V}\rangle$ consists of the following three main elements.

\textbf{1.} Causal graph $\G = (\V, \mathcal{E})$ is a directed acyclic graph (DAG) on a set of vertices $\V$ with the set of edges $\mathcal{E}$. The set of vertices $\V = \N \cup \U$ consists of the set of all random variables in the system, where $\N$ is the set of observed random variables and $\U$ is a set of latent variables. The set of edges $\mathcal{E}$ represents the causal dependencies between the vertices. For each edge $(X, Y) \in \mathcal{E}$, we say that $X$ is a parent of $Y$ or, equivalently, $Y$ is a child of $X$. For arbitrary $X, Y \in\V$, the vertex $X$ is an ancestor of $Y$ and $Y$ is a descendant of $X$ if there is a directed path from $X$ to $Y$ in graph $\G$. Note that both $X$ and $Y$ are ancestors and descendants of themselves. Let $\Pa{X}{\G}$, $\Ch{X}{\G}$ denote the set of parents and children of vertex $X$, respectively. With a slight abuse of notation, let $\Anc{X}{\G}$ and $\De{X}{\G}$ denote the set of ancestors and descendants of $\X$ that belong to $\N$ in graph $\G$. The aforementioned definitions can be extended to sets of vertices via a union operation, e.g. $\Pa{\X}{\G} = \bigcup_{X \in \X}\Pa{X}{\G}$. For two disjoint subsets $\X, \Y \subset \V$,  let $\G_{\overline{\X}, \underline{\Y}}$ denote the subgraph of $\G$ obtained by removing all incoming edges into $\X$ and all outgoing edges from $\Y$.
% \begin{definition}[Blocked path \cite{pearl2009causality}]
% A path $p$ in $\G$ is a non-repeated sequence of connected nodes and is said to be blocked by a set of nodes in $\Z$ if and only if\\
% $\bullet$ $p$ contains a chain $X\rightarrow W \rightarrow Y$ or fork $X\leftarrow W \rightarrow Y$, such that $W \in \Z$, or\\
% $\bullet$ $p$ contains a collider $X \rightarrow W \leftarrow Y$ (node $W$ is called a collider), such that $\Z \cap \De{W}{\G}=\emptyset$.
% \end{definition}
Also, disjoint sets $\X$ and $\Y$ are \textit{d-separated} by $\Z$ in $\G$ if all paths between $\X$ and $\Y$ are blocked\footnote{See definition in \cite{pearl2009causality}.} by $\Z$, which is denoted by $\X \independent_{\G} \Y|\Z$. Given the \textit{global Markov property}, $ \X \independent_{\G} \Y|\Z$ implies $ \X \independent \Y|\Z$. 
\\
We further assume that the graph $\G$ corresponding to SCM $\M$ is semi-Markovian, which means that any vertex $U\in \U$ has exactly two children and zero parents. This assumption is common in the literature, since any SCM $\M$ has a semi-Markovian projection \citep{verma1993graphical} that preserves all causal relationships among the observed variables. Note that our results would be valid for general SCM models. Furthermore, let $\G_\X$ denote the induced subgraph of $\G$ by a set of variables $\X$ and unobserved variables with both children in $\X$. We say that $\X \subseteq \N$ is an ancestral set in $\G$, if $\X = \Anc{\X}{\G}$.

\textbf{2.} In SCM $\M$ the exogenous noises $\{\epsilon_V\}_{V \in \V}$ are mutually independent random variables, and each random variable $\epsilon_{V}$ corresponds to the exogenous noise of a random variable $V$. 

\textbf{3.} The set of functions $\{f_{V}(\cdot)\}_{V\in \V}$ defines the relationship between the random variables in the causal graph $\G$ through the structural equation $V = f_{V}(\Pa{V}{\G}, \epsilon_V)$ for any $V\in \V$. In a causal probabilistic model, the post-interventional distribution is defined by $do$-operation. Let $\X \subseteq \V$ and $\x \in \dom{\X}{}$, then the intervention $do(\X=\x)$ transforms $\M$ to a new SCM in which all the structural equations of $\X$ are replaced by $\X = \x$ and graph $\G$ is replaced by $\G_{\overline{\X}}$. 
% This operation also transforms the causal graph $\G$ to a new graph $\G_{\overline{\X}}$, as each random variable $X \in \X$ does not depend on any other random variable in $\V$.
\\
For the sake of simplicity, we further assume that all random variables are \textit{discrete}. Under such an assumption, a probabilistic model $\M$ may be equivalently defined by the causal graph $\G$ and by the set of conditional distributions $\{P^{\M}(V|\Pa{V}{\G})\}_{V\in \V}$ instead of $\{f_{V}(\cdot)\}$ and $\{\epsilon_V\}_{V \in \V}$. By the \textit{Markov factorization property}, this set of conditional distributions induces a joint distribution over the random variables $\V$ as follows $P^{\M}(\mathbf{V}) = \prod_{V\in \V}P^{\M}(V|\Pa{V}{\G})$.
% \begin{equation*}
%     P^{\M}(\mathbf{V}) = \prod_{V\in \V}P^{\M}(V|\Pa{V}{\G}).
% \end{equation*}
For notational simplicity, we use $do(\x)$ instead of $do(\X=\x)$ to denote an intervention, e.g. $P(y|do(x)):= P(y|do(X=x))$. 

Finally, we use $\mathbb{M}(\G)$ to denote the set of all possible SCMs with causal graph $\G$ such that 
$P^{\M}(\n) > 0$ for any $\M \in \mathbb{M}(\G)$ and arbitrary realization $\n \in \dom{\N}{}$. This is known as the \textit{positivity assumption} which is crucial in the identifiability literature \citep{kivva2022revisiting}. 

\subsection{Universal identifiability problem}
Before presenting a formal definition of completeness, we define a general form of identifiability problem. 
\begin{definition}
    \label{def:general form}
    Let a \textbf{query} $\mathcal{Q}=(\G,\mathcal{P})$ specify a causal graph $\G$ and a family $\mathcal{P}=\{P(\A_i|do(\B_i),\mathbf{C}_i)\}_{i=0}^{m}$, where $(\A_i,\B_i,\mathbf{C}_i)$ is a triple of disjoint subsets of $\N$. For $\M\in\mathbb{M}(\G)$, write $\mathcal{Q}^{\M}=(\G,\mathcal{P}^{\M})$ for the corresponding available distributions evaluated in $\M$. Given disjoint sets $\X,\Z,\Y\subseteq\N$, the causal effect $P(\y|do(\x),\z)$ is \textbf{identifiable} from $\mathcal{Q}$ if any two arbitrary models $\M_1,\M_2\in\mathbb{M}(\G)$ with $\mathcal{Q}^{\M_1}=\mathcal{Q}^{\M_2}$ satisfy $P^{\M_1}(\y|do(\x),\z)=P^{\M_2}(\y|do(\x),\z)$ for every realization.
\end{definition}
%Next, we define the Universal Identifiability Problem.
\begin{definition}[Universal Identifiability Problem (uID)]\label{def:uID}
    Let $\G$ be a given causal graph, and query $\mathcal{Q}=(\G, \mathcal{P})$ is given such that $\mathcal{P}=\{P(\A_i|do(\B_i))\}_{i=0}^{m}$, where $\A_i\cap \B_i=\emptyset$ and $\A_i, \B_i \subseteq \N$ for $i=0,\dots,m$. Then, for disjoint sets $\X, \Y\subseteq\N$, the causal effect $P(\y|do(\x))$ is universally identifiable (uID) from $\mathcal{Q}$ if $P^{\M}(\y|do(\x))$ is uniquely recoverable from $\mathcal{Q}^{\M}$, in the sense of Definition~\ref{def:general form}, in arbitrary causal model $\M \in \mathbb{M}(\G)$.
\end{definition}

% \begin{definition}[Universal Identifiability Problem (uID)]\label{def:uID}
%     Let $\G$ be a given causal graph, and suppose $\X$ and $\Y$ are disjoint subsets of $\N$. Moreover, let $\mathbb{A} = \{\A_i\}_{i=0}^{m}$ and $\mathbb{B} = \{\B_i\}_{i=0}^{m}$ be collections of subsets of $\N$, such that $\A_i\cap \B_i$ for all $i\in 0\dots m$.  Then the causal effect $P(\y|do(\x))$ is universal-identifiable (uID) from $(\mathbb{A}, \mathbb{B})$ in $\G$ if $P^{\M}(\y|do(\x))$ is uniquely computable from distributions $\{P^{\M}\left(\A_i|do(\B_i)\right)\}_{i=0}^{m}$ in any causal model $\M \in \mathbb{M}(\G)$.
% \end{definition}
Note that classical ID or gID are special cases of uID problem, where query $\mathcal{Q}$ equals $(\G, P(\N))$ for ID problem and  $(\G, \{P(\A_i|do(\N\setminus\A_i))\}_{i=0}^{m})$ for gID problem. We shall prove that $do$-calculus is \textbf{not complete} for uID after first providing a definition of completeness for the aforementioned general form of identification problem.

\subsection{Formal definition of completeness}
\label{sec: def do-inference}
In this section, we provide a formal definition for the completeness of $do$-calculus rules for the identifiability problem that is in the general form of Definition \ref{def:general form}. Note that classical ID, gID, conditional gID (c-gID) \citep{kivva2023identifiability}, and uID problems are specific forms of Definition \ref{def:general form}, so our definition of completeness applies to those problems as well. Before introducing our formal definition, we need to review some preliminaries.

 The rules of $do$-calculus were introduced by \cite{pearl1995causal} and defined as follows. Assume $\X, \Y, \W$ and $\Z$ are disjoint subsets of $\V$.\\
 \textbf{DR1}. \textbf{Insertion/deletion of observations}. If $\Y\independent_{\G_{\overline{\X}}}\Z | \X, \W$:
 \begin{equation*}
    P(\y|do(\x), \z, \w) = P(\y|do(\x), \w).
\end{equation*}
\textbf{DR2}. \textbf{Action/observation exchange}. If $\Y \independent_{\G_{\overline{\X},\underline{\Z}}} \Z | \X, \W$:
% $\Y$ and $\Z$ are d-separated by $\X$ and $\W$ in a graph where incoming edges in $\X$ and outgoing edges from $\Z$ have been removed:
\begin{equation*}
    P(\y|do(\x), do(\z), \w) = P(\y|do(\x), \z, \w).
\end{equation*}\\
\textbf{DR3}. \textbf{Insertion/deletion of actions}. If $\Y \independent_{\G_{\overline{\X},\overline{\Z(\W)}}} \Z | \X, \W$,
% $\Y$ and $\Z$ are d-separated by $\X$ and $\W$ in a graph where incoming edges in $\X$ and $\Z(\W)$ have been removed. $\Z(\W)$ is the subset of nodes in $\Z$ that are not ancestors of any node in $\W$ in a graph that is obtained from $\G$ after removing all edges into $\X$:
\begin{equation*}
    P(\y|do(\x), do(\z), \w) = P(\y|do(\x), \w),
\end{equation*}
where $\Z(\W)$ is the subset of nodes in $\Z$ that are not ancestors of any node in $\W$ in $\G_{\overline{\X}}$.

In general, for the identification of the causal effect, there may exist some function $f(\cdot)$ that may compute the target causal effect given the corresponding input, but there is no guarantee that this function can be "well expressed". A classical example is the Abel--Ruffini theorem \citep{abel1824memoire}: there is no general formula in radicals for the roots of a polynomial of degree five or higher. Similarly, there may be no guarantees that the $do$-calculus is enough in terms of identifying the function $f(\cdot)$. 
% But before answering whether $do$-calculus is complete for the given problem, we must clarify and formally define what it means that $do$-calculus is complete. 

Intuitively, the statement "$do$-calculus is complete" means that these rules provide a \textit{complete} set of \textit{inference rules} that suffice to identify an identifiable target causal effect, i.e. they generate \textbf{all possible algebraic transformations} needed to express identifiable causal effects. Therefore, we would only fail to find an expression for a causal effect in terms of available distributions via do-calculus rules when the effect is unidentifiable. 

We will allow the use of the standard probability manipulations defined below along with the "$do$-calculus" rules.

\textbf{Standard probability manipulation rules.}\label{PR rules}
Suppose $\Y$, $\X$, $\W$ and $\Z$ are disjoint subsets of $\V$.\\
\textbf{PR1. Complement rule}. $P(\y_i|do(\x),\w) = 1 - \sum_{\y \in \dom{\Y}{} \setminus \{\y_i\}} P(\y|do(\x), \w)$.
% \begin{equation*}
%     P(\y_i|do(\x),\w) = 1 - \sum_{\y \in \dom{\Y}{} \setminus \{\y_i\}} P(\y|do(\x), \w)
% \end{equation*}
\\
\textbf{PR2. Marginalization}. $P(\y|do(\x), \w) = \sum_{\z \in \dom{\Z}{}} P(\y, \z|do(\x), \w)$.
% \begin{equation*}
%     P(\y|do(\x), \w) = \sum_{\z \in \dom{\Z}{}} P(\y, \z|do(\x), \w).
% \end{equation*}
\\
\textbf{PR3. Chain rule}. $P(\y, \z|do(\x), \w) = P(\y|do(\x), \z, \w) P(\z|do(\x), \w)$.
% \begin{equation*}
%     P(\y, \z|do(\x), \w) = P(\y|do(\x), \z, \w) P(\z|do(\x), \w) 
% \end{equation*}
\\
\textbf{PR4. Conditioning}. $P(\y|do(\x),\z, \w) = \frac{P(\y, \z|do(\x), \w)}{P(\z|do(\x), \w)}$.
% \begin{equation*}
%     P(\y|do(\x),\z, \w) = \frac{P(\y, \z|do(\x), \w)}{P(\z|do(\x), \w)}
% \end{equation*}
% \\
% 5. \textbf{Independence}. $\X$, $\Y$, $\Z$ are disjoint subsets of random variables. If $\Y \independent \X | \Z$, then $P(\y|\x, \z) = P(\y|\z)$
% \\
% 6. \textbf{Global Markov property}.
% \begin{equation*}
%     \Y \independent_{\G} \X | \Z \Rightarrow \Y \independent \X | \Z
% \end{equation*}
% for all disjoint vertex sets $\Y$, $\X$, $\Z$.
\\Note that the first rule of $do$-calculus is an extension of the independence rule (i.e. $P(\y|\x, \z) = P(\y|\z)$) of standard probability manipulation, that is why we do not include the independence rule in the list above.\\
\textbf{Remark}. As a consequence of $PR2$, we have $1 = \sum_{\y \in \dom{\Y}{}}P(\y|do(\x), \w)$.
% \begin{equation*}
%     1 = \sum_{\y \in \dom{\Y}{}}P(\y|do(\x), \w).
% \end{equation*}

From here on, we call the standard probability manipulation rules together with the rules of $do$-calculus: the $\textbf{\textit{do}}$-\textbf{inference} rules. We are now ready to propose a formal definition for the completeness of $do$-calculus.
\begin{definition}\label{def:doID}
    Suppose $\Z, \X, \Y \subseteq \N$ are three disjoint sets. The interventional probability $P(\y|do(\x), \z)$ is $\textbf{do}$-\textbf{identifiable} given the query  $\mathcal{Q}=(\G, \mathcal{P})$ \footnote{See Definition  \ref{def:general form}.} if there is a sequence of $do$-inference rules that transforms $P(\y|do(\x), \z)$ into an expression involving only distributions in $\mathcal{Q}$.
\end{definition}
\begin{definition}[Completeness of $do$-calculus]\label{def:do-calculus complete}
     $Do$-calculus is \textbf{complete} for a specific identifiability problem if every causal effect that is identifiable for this identifiability problem in terms of Definition~\ref{def:general form} is also $do$-identifiable.
\end{definition}
\textbf{Remark.} In addition to $do$-inference rules, we allow the use of standard arithmetic manipulations such as associativity, the distributive law, and factoring out common terms. \footnote{Note that we do not allow operations that introduce new quantities into the expression at any stage and just allow for sequential application of $\textbf{\textit{do}}$-\textbf{inference} rules to the terms in the existing expression.} %However, adding and subtracting the same quantity within an expression is not permitted, as we do not consider this a standard arithmetic manipulation. Likewise, multiplication and division by the same quantity are also disallowed. The primary reason for these restrictions is that, in general, such operations may introduce new quantities into the expression that does not have a ``probabilistic interpretation".

% \begin{wrapfigure}{R}{0.5\textwidth}
% \vspace{-6ex}
% \hspace{-5ex}
% \begin{minipage}{0.5\textwidth}
%     \textbf{Input:} $\G$, c-component $S$ in $\G_{\A}$, $Q[\A](\cdot)$
    
%     \begin{algorithmic}[1]
%     \STATE $\A_0 \gets \A$
%     \STATE $\S_0 \gets$ c-component in $\G_{\A_0}$ that contains $\S$ 
%     \STATE $k \gets 0$
%     \WHILE{$\S_k\neq \S$ and $\S_k\neq\S_{k-1}$}
%     \STATE $k \gets k+1$
%     \STATE $\A_k \gets \Anc{\S}{\G_{\S_{k-1}}}$
%     \STATE $\S_k \gets$ c-component in $\G_{\A_k}$ that contains $\S$
%     \ENDWHILE
%     \IF{$\S_k = \S_{k-1}$}
%     \STATE \textbf{Return:} False
%     \ENDIF
%     \STATE \textbf{Return:} True
%     \end{algorithmic}
%     \captionsetup{type=algorithm}
%     % \caption{Identification of $Q[\S]$ from $(Q[\A], \G)$}
%     \caption{$ID\_Single(\S, \G_{\A})$}
%     \label{alg:compute Q}
% \end{minipage}
% \vspace{-3ex}
% \end{wrapfigure}

% \begin{algorithm}[t]
%     \caption{Identification of $Q[\S]$ from $(Q[\A], \G)$}
%     \label{alg:compute Q}
%     \textbf{Input:} $\G$, c-component $S$ in $\G_{\A}$, $Q[\A](\cdot)$
    
%     \begin{algorithmic}[1]
%     \STATE $\A_0 \gets \A$
%     \STATE $\S_0 \gets$ c-component in $\G_{\A_0}$ that contains $\S$ 
%     \STATE $k \gets 0$
%     \WHILE{$\S_k\neq \S$ and $\S_k\neq\S_{k-1}$}
%     \STATE $k \gets k+1$
%     \STATE $\A_k \gets \Anc{\S}{\G_{\S_{k-1}}}$
%     \STATE $\S_k \gets$ c-component in $\G_{\A_k}$ that contains $\S$
%     \ENDWHILE
%     \IF{$\S_k = \S_{k-1}$}
%     \STATE \textbf{Return:} False
%     \ENDIF
%     \STATE \textbf{Return:} True
%     \end{algorithmic}
% \end{algorithm}

\section{\texorpdfstring{$Do$}{Do}-calculus is complete in ID, gID and c-gID  problems}
The problem of identifiability has been extensively studied in the literature under various assumptions.
% ranging from restrictions on the causal graph or the structure of observational data to more complex settings involving missing data or multiple environments. 
In this section, we focus on the classic line of identifiability problems initiated by \cite{tian2003ID} and continued by \cite{shpitser2006identification, shpitser2012identification, huang2006identifiability, huang2008completeness, lee2019general, kivva2022revisiting, kivva2023identifiability}. All aforementioned identifiability results rely on a graphical criterion based on the notion of \textit{c-component}, the definition of which can be found in Appendix \ref{app:proofs sec 3}. Note that these works are special cases of c-gID problem studied in \cite{kivva2023identifiability}.
\label{sec:alignment with previous works}
% Thus, it is enough to revisit the identification algorithms presented by \cite{shpitser2006identification, kivva2022revisiting, kivva2023identifiability} under the proposed formal definition of completeness of $do$-calculus in Definition \ref{def:do-calculus complete} and show that it agrees with the previous results.

More specifically, \cite{shpitser2006identification} proposed a graphical criterion to solve the ID problem, that is, the conditions under which the causal effect $P(\y|do(\x))$ of interest can be recovered uniquely given the causal graph $\G$ and the observational distribution $P(\N)$. When the conditions are satisfied, the authors provided an algorithm that computes $P(\y|do(\x))$ based on $do$-calculus rules and probability manipulations; otherwise, the causal effect cannot be recovered uniquely. As a corollary, they concluded that the rules of $do$-calculus together with standard probability manipulations are sound and complete for the classical ID problem. With a similar procedure, this result was extended for the problems of gID and c-gID \citep{lee2019general, kivva2022revisiting, kivva2023identifiability}. It should be noted that these completeness results are restricted by the contexts of ID, gID, and c-gID problems and the specific algorithms proposed there. %For other variations of identifiability problems, a separate analysis is required.

Although for the aforementioned  problems, the completeness of the $do$-calculus through analysis of a specific algorithm has been established, to the best of our knowledge,  no formal definition for the completeness of $do$-calculus rules prior to this work has been proposed in the literature. Therefore, it is important to verify that our definition of completeness of $do$-calculus (Definition \ref{def:do-calculus complete}) is well aligned with the previous results \citep{shpitser2006identification, shpitser2012identification, lee2019general, kivva2022revisiting, kivva2023identifiability} as summarized in the following theorem. The proof of the theorem is in Appendix \ref{app:proofs sec 3}.
\begin{restatable}{theorem}{thDoCalculusWellDefined}\label{thm_completeness for gID}
    $Do$-calculus is complete in terms of Definition \ref{def:do-calculus complete} for the ID, gID and c-gID problems studied in \citep{shpitser2006identification}, \citep{kivva2022revisiting}, and \citep{kivva2023identifiability}, respectively.
\end{restatable}
 %This theorem concludes that our Definition \ref{def:do-calculus complete} is consistent with the previous understanding of completeness of $do$-calculus rules.

\section{Main Result: \texorpdfstring{$do$}{do}-calculus is not complete for uID}
\label{sec:main result}
\setlength{\intextsep}{0pt}
\setlength{\columnsep}{5pt}
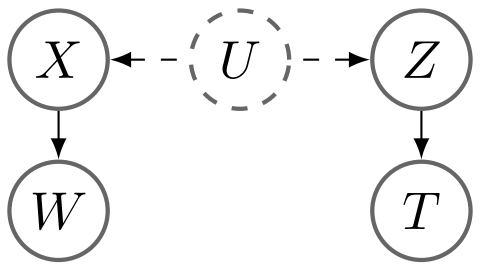
\begin{wrapfigure}{r}{0.5\textwidth}
\centering
\begin{tikzpicture}[
        roundnode/.style={circle, draw=black!60,, fill=white, thick, inner sep=1pt, minimum size=0.65cm},
        dashednode/.style = {circle, draw=black!60, dashed, fill=white, thick, inner sep=1pt, minimum size=0.65cm},
        ]
            % Nodes
            \node[roundnode]        (W)        at (-1.2, 0)                  {$W$};
            \node[roundnode]        (T)        at (1.2, 0)                  {$T$};
            \node[roundnode]        (X)        at (-1.2, 1)               {$X$};
            \node[roundnode]        (Z)        at (1.2, 1)                {$Z$};
            \node[dashednode]        (U)        at (0, 1)                    {$U$};
            
            %Edges
            \draw[latex-, dashed] (X) -- (U);
            \draw[latex-, dashed] (Z) -- (U);
            \draw[latex-] (W) -- (X);
            \draw[latex-] (T) -- (Z);
            % \draw[-latex] (X1.south) -- (Z1.north);
            % \draw[latex-] (Z1.east) -- (Z2.west);
            % \draw[latex-] (Z2.east) -- (W1.west);
            % \draw[latex-] (W1.north) -- (Y1.south);
            % \draw[latex-] (Z1.west) -- (U1);
            % \draw[latex-] (X1.west) -- (U1);
            % \draw[latex-] (Y1.east) -- (U2);
            % \draw[latex-] (W1.east) -- (U2);
            % \draw[latex-] (Z2) -- (X1);
            % \draw[latex-latex, dashed] (X1) .. controls +(left:10mm ) and +(left:10mm) .. (Z1);
            % \draw[latex-latex, dashed] (Z1.north east) .. controls +(up:5mm) and +(up:5mm) .. (W1.north west);
            % \draw[latex-latex, dashed] (W1) .. controls +(right:10mm ) and +(right:10mm) .. (Y1);
        \end{tikzpicture}
\caption{DAG $\mathcal{G}$ with $\mathbf{N} = \{X, W, Z, T\}$ and $\mathbf{U} = \{U\}$.}
   
    \label{fig: main graph}
    % \vspace{-3ex}
\end{wrapfigure}
% Up until now, we have observed that $do$-calculus is complete for the studied identifiability problems. 
In this section, we show that $do$-calculus is not complete for the  uID problem, defined in Definition \ref{def:uID}. We do so by providing an instance of uID, where a causal effect can be uniquely identified (see Definition \ref{def:uID}) but not through application of $do$-inference rules as required by Definition \ref{def:doID}.
%\begin{definition}[Universal Identifiability Problem (uID)]
  %  Suppose $\X$, and $\Y$ are disjoint subsets of $\N$, and let $\G$ be a given causal graph. Additionally, $\mathbb{A} = \{\A_i\}_{i=0}^{m}$ and $\mathbb{B} = \{\B_i\}_{i=0}^{m}$ are collections of subsets of $\N$, such that $\A_i\cap \B_i$ for all $i\in 0\dots m$.  Then the causal effect $P(\y|do(\x))$ is said to be universal-identifiable (uID) from $(\mathbb{A}, \mathbb{B})$ in $\G$ if $P^{\M}(\y|do(\x))$ is uniquely computable from observational distributions $\{P^{\M}\left(\A_i|do(\B_i)\right)\}_{i=0}^{m}$ in any causal model $\M \in \mathbb{M}(\G)$.
%\end{definition}
%As classical ID or gID are partial cases of uID problem. Let us further propose a specific setting of uID problem which we will use to prove that $do$-calculus is \textbf{not complete}. 
% Such a result is strong evidence of the complexity of identifiability problems going beyond the settings of gID. 
%\subsection{

\textbf{$\mathbf{Prob_{uID}}$: (An instance of uID problem)} Consider the causal graph $\G$ in Figure~\ref{fig: main graph} and let query $\mathcal{Q}^{\mathcal{M}} = (\G, \{P^{\mathcal{M}}(\A_i| do(\B_i))\}_{i=0}^{2})$ be such that $\A_0 = \{X, Z\}, \A_1 = \{X, T\}, \A_2 = \{Z, W\}$ and $\B_0 = \B_1 = \B_2 = \emptyset$. We would like to know whether $P^{\M}(\y)$ is uID from $\mathcal{Q}^{\M}$ for $\Y=\{W, T\}$ in any causal model $\M \in \mathbb{M}(\G)$.

In other words, we would like to know whether in Figure~\ref{fig: main graph}, $P^\M(w, t)$ can be uniquely computed from the  distributions: $P^{\M}(X, Z)$, $P^{\M}(X, T)$ and $P^{\M}(Z, W)$, in any causal model $\M \in \mathbb{M}(\G)$.  

% We show that $P(Y,T)$ is identifiable in general sense, i.e. $P^{\mathcal{M}}(Y,T)$ is uniquely computable from $P^{\mathcal{M}}(X,Z)$, $P^{\mathcal{M}}(X,T)$ and $P^{\mathcal{M}}(Y,Z)$ for any SCM $\mathcal{M}$. But at the same time it is not identifiable in a sense of Definition~\ref{def: identifiability}.

\subsection{Identifiability of \texorpdfstring{$\mathbf{Prob_{uID}}$}{Prob\_uID} }
\label{sec:identifiability}
In this section, we show that in Figure~\ref{fig: main graph}, $P^{\mathcal{M}}(w,t)$ is uID, that is, given the distributions $P^{\M}(X, Z)$, $P^{\M}(X, T)$ and $P^{\M}(Z, W)$, it takes a unique value. 

\begin{theorem}
    \label{th:main:identifiable} In $\mathbf{Prob_{uID}}$,
    $P^{\mathcal{M}}(\y):=P^{\mathcal{M}}(w, t)$ is uID  given the query $\mathcal{Q}$.
\end{theorem}
\begin{proof}
Let $\mathcal{M}_1, \mathcal{M}_2 \in \mathbb{M}(\G)$ be two SCMs with $\mathcal{Q}^{\M_1} = \mathcal{Q}^{\M_2}$, and fix an arbitrary realization $(w, t)$. We show that $P^{\M_1}(w, t) = P^{\M_2}(w, t)$.

Recall that we assumed that all random variables are discrete. Hence for $\M \in \{\M_1, \M_2\}$ and the fixed $(w, t)$ we have
\begin{gather*}
    P^{\mathcal{M}}(x_i, z_j)=c^{\mathcal{M}}_{ij}, P^{\mathcal{M}}(t|z_j) = a^{\mathcal{M}}_j,  P^{\mathcal{M}}(w|x_i)=b^{\mathcal{M}}_i, \\ P^{\mathcal{M}}(z_j, w) = d^{\mathcal{M}}_j, P^{\mathcal{M}}(x_i, t) = e^{\mathcal{M}}_i.
\end{gather*}
 We can then define matrix $C^{\mathcal{M}}$ and vectors $\mathbf{a}^{\mathcal{M}}$, $\mathbf{b}^{\mathcal{M}}$, $\mathbf{d}^{\mathcal{M}}$, $\mathbf{e}^{\mathcal{M}}$ whose respective entries are defined above. Note that the matrices $C^{\M_k}$ and vectors $\d^{\M_k}, \e^{\M_k}$ are known from query $\mathcal{Q}^{\M_k}$ and are equal for  SCMs $\M_1$ and $\M_2$. For ease of notation, we will drop the superscripts denoting SCM when there is no ambiguity. Then,
 % Next, we establish important relationships for the probabilities $P(z_j, w)$, $P(x_i, t)$ and $P(w, t)$ in graph $\G$. 
 %\begin{equation*}
     $P(z_j, w) = \sum_{i}P(x_i, z_j, w) \overset{(*)}{=} \sum_{i}P(w|x_i)P(x_i, z_j).$
 %\end{equation*}
In $(*)$ we used the independence relation $W \independent Z | X$ since $W$ and $Z$ are d-separated by $X$ in $\G$. A similar equation can be obtained for $P(x_i, t)$ due to the symmetry of the graph $\G$. Note that these equations can be written in the matrix form as $\d = \C^T\b^{\M_k}$ and $\e = \C \a^{\M_k}$ corresponding to $P^{\M_k}(z_j, w)$ and $P^{\M_k}(x_i, t)$ respectively. On the other hand, $P(w, t) = \sum_{i} P(t, x_i, w) \overset{(*)}{=} \sum_i P(w|x_i)P(x_i, t)$,
% \begin{align}
%     & P(w, t) = \sum_{i} P(t, x_i, w) \overset{(*)}{=} \sum_i P(w|x_i)P(x_i, t),
% \end{align}
where $(*)$ holds because $W\independent T|X$.
Equivalently, in the matrix form, we have  $P^{\M_k}(w, t) = \e^T\b^{\M_k}$. Thus, we have
\begin{align*}
    & \e^T\b^{\M_1} = \left(\C\a^{\M_{1}}\right)^T\b^{\M_1} = \left(\a^{\M_{1}}\right)^T\C^T\b^{\M_1} \\
    & \overset{(*)}{=} \left(\a^{\M_{1}}\right)^T\C^T\b^{\M_2} \overset{(**)}{=} \left(\a^{\M_{2}}\right)^T\C^T\b^{\M_2} = \e^T\b^{\M_2},
\end{align*}
 where $(*)$ is based on the equality $\d = \C^T\b^{\M_k}$ and $(**)$ is based on the equality $\e = \C \a^{\M_k}$. This implies that $P^{\M_1}(w, t) = P^{\M_2}(w, t)$. Since $(w, t)$ was arbitrary, $P(\y)$ is uID given $\mathcal{Q}$.
\end{proof}

As we saw, in order to identify the causal effect $P(w, t)$, we had to find a solution $\b^*$ of the linear equation $\d = \C^T\b$ (a solution always exists, namely the true $\b$) and use it to compute $P(w, t)$ through equality $P(w, t) = \e^T\b^*$.
% \begin{equation*}
%     P(w, t) = \e^T\b^*.
% \end{equation*}
Note that the linear equation $\d = \C^T\b$ with parameter $\b$ may have infinitely many solutions, but for the purpose of the proposed identification procedure, it does not matter which solutions we choose. 
The takeaway lesson from this example is that there is no meaningful interpretation in terms of probability manipulations for the process of finding $\b^*$. This suggests that $do$-calculus rules may not be complete for the uID. 
% Simultaneously, this illustrates that the structure complexity of uID problem is much higher than the problems studied in literature before.

To have a complete picture of the problem in Figure~\ref{fig: main graph}, we show the following non-identifiability result, proof of which can be found in Appendix~\ref{app:proofs sec 4}. 

\begin{proposition}
    \label{prop:sec4:non-id}
    In $\mathbf{Prob_{uID}}$ the causal effects $P^{\mathcal{M}}(T| Z)$, $P^{\mathcal{M}}(W| X)$ and $P^{\mathcal{M}}(X, W, Z, T)$ are not identifiable from $\mathcal{Q}$ in terms of Definition~\ref{def:general form}.
\end{proposition}

\subsection{Do-calculus is not complete }
\label{sec:do-calculus is not complete}
In this section, we show that $do$-calculus rules are not complete for uID problem. We do so using our example problem, $\mathbf{Prob_{uID}}$. As established in Theorem \ref{th:main:identifiable}, $P(w, t)$ is uID \footnote{ Recall that in the context of our example $\mathbf{Prob_{uID}}$, we mean that $P(w, t)$ is uniquely computable from the query $\mathcal{Q}=(\G, \{P(X, Z), P(Z, W), P(X, T)\})$.}, but as we shall prove it is not $do$-identifiable from $\mathcal{Q}$ as defined in Definition \ref{def:doID}. 

To prove it we will consider a specific setup of $\mathbf{Prob_{uID}}$ for which we prove the following theorem.
\begin{theorem}\label{th:main: prob-uid not do-identifiable}
Consider the causal graph $\mathcal{G}$ of Figure~\ref{fig: main graph} with binary
observed variables $X, Z, W, T \in \{-1,1\}$, and let $\mathbb{M}(\mathcal{G})$
denote the set of positive SCMs compatible with $\mathcal{G}$. Fix any
$(w,t) \in \{-1,1\}^2$. There is no rational function
$\mathcal{F}_{w,t} : \mathbb{R}^{12} \to \mathbb{R}$ such that, for every
$\mathcal{M} \in \mathbb{M}(\mathcal{G})$, $\mathcal{F}_{w,t}$ is well defined at
$\bigl(P^{\mathcal{M}}(X,Z),\, P^{\mathcal{M}}(X,T),\, P^{\mathcal{M}}(Z,W)\bigr)$ and
\[
  P^{\mathcal{M}}(W=w, T=t) \;=\; \mathcal{F}_{w,t}\bigl(P^{\mathcal{M}}(X,Z),\, P^{\mathcal{M}}(X,T),\, P^{\mathcal{M}}(Z,W)\bigr),
\]
where each distribution is viewed as the vector of its four probabilities.
\end{theorem}
\begin{proof}
Let $X,Z,W,T$ have alphabet $\{-1,1\}$ and let $U$ have alphabet $\mathcal{X}(U)=\{-1,1\}^2$, with elements written $(i,j)$. Without loss of generality we will prove the statement of the theorem for the $w=1$ and $t=1$.

Define the SCM $\M$ via the set of conditional probabilities:
    \begin{equation}
    \begin{split}
        \label{eq: SCM M}
        \centering
        P\left(U=\left(i, j\right)\right) &= \frac{1+4cij}{4},\\
        P\left(X=x|U=\left(i, j\right)\right) = \frac{1+xi/2}{2},\; &\; P\left(Z=z|U=\left(i, j\right)\right) = \frac{1 + zj/2}{2}, \\
        P\left(W=w|X=x\right) = \frac{1 + axw}{2}, \; & \; P\left(T=t|Z=z\right) = \frac{1+bzt}{2},
    \end{split}
    \end{equation}
    where $|c| < 1/4$, $|a| < 1$ and $|b| < 1$. Note that all these conditional probabilities are positive and sum to 1 over all realizations. Therefore, they define a valid SCM $\M$ which satisfies a positivity assumption.

    From the Markov factorization property, we have the following:
    \begin{equation*}
    \begin{split}
        P(x, z, w, t) &= \sum_{u \in \dom{U}{}} P(u)P(w|x)P(x|u)P(z|u)P(t|z) \\
        &= \sum_{i, j \in \{-1, 1\}} \frac{1+4cij}{4} \frac{1+\frac{xi}{2}}{2} \frac{1 + \frac{zj}{2}}{2} \frac{1 + axw}{2} \frac{1+bzt}{2} \\
        & = \frac{4(1 + cxz)}{16} \frac{(1+axw)(1+bzt)}{4} = \frac{(1 + cxz)(1+axw)(1+bzt)}{16}.
    \end{split}
    \end{equation*}
    Consequently, from the joint $P(x, z, w, t)$, we can obtain by marginalization:
    \begin{align}
        & P(x, z) = \frac{1+cxz}{4},\\
        & P(x, t) = \frac{1+cbxt}{4}, \\
        & P(z, w) = \frac{1+cazw}{4}, \\
        \label{eq:5}
        & P(w, t) = \frac{1+cabwt}{4}.
    \end{align}

    Denote $\alpha_1 = cb$ and $\alpha_2 = ca$. Since $x, z, w, t \in \{-1,1\}$, for each realization the observed probabilities $P(x,z)$, $P(x,t)$ and $P(z,w)$ are affine (in particular, polynomial) functions of $(c, \alpha_1, \alpha_2)$.% while $P(w,t) = \tfrac14\bigl(1 + \tfrac{\alpha_1\alpha_2}{c}\,wt\bigr)$ is not.

    Suppose, for contradiction, that a rational function  $\F_{w, t}$ exists for each $(w, t)$. Let $R(c,\alpha_1,\alpha_2)$ denote the vector of observed probabilities $P(X,Z)$, $P(X,T)$, $P(Z,W)$, whose entries are affine in $(c,\alpha_1,\alpha_2)$. Since a rational function of affine functions is again rational, there exist multi-variable polynomials $R_1(\cdot), R_2(\cdot)$ such that
    \begin{equation*}
        P(W=1, T=1) = \F_{1, 1}(R(c, \alpha_1, \alpha_2)) = \frac{R_{1}(c, \alpha_1, \alpha_2)}{R_2(c, \alpha_1, \alpha_2)}.
    \end{equation*}
    %where $R_1(\cdot), R_2(\cdot)$ are multi-variable polynomials and $R(c, \alpha_1, \alpha_2)$ is a vector-valued function that maps into a polynomial representation of the realizations of $P(X, Z)$, $P(X, T)$, $P(Z, T)$. 

    Every $(c,\alpha_1,\alpha_2)$ in the nonempty open set
\[
  \mathcal{O} = \{(c,\alpha_1,\alpha_2) : 0 < c < 1/4,\ |\alpha_1| < c,\ |\alpha_2| < c\}
\]
corresponds to a model in~\eqref{eq: SCM M} (take $a = \alpha_2/c$ and $b = \alpha_1/c$, so that $|a|,|b| < 1$). Moreover, the point $(0,0,0)$ corresponds to the models with $c = 0$. Since $\mathcal{F}_{1,1}$ is well defined at every model, we have $R_2 \neq 0$ on $\mathcal{O}$ and $R_2(0,0,0) \neq 0$.

By~\eqref{eq:5}, $P(W=1,T=1) = \tfrac14(1 + cab)$, and for $c \neq 0$ we have $cab = (ca)(cb)/c = \alpha_1\alpha_2/c$. Hence, on $\mathcal{O}$,
\[
  \frac{\alpha_1\alpha_2}{c} \;=\; 4\,\frac{R_1(c,\alpha_1,\alpha_2)}{R_2(c,\alpha_1,\alpha_2)} - 1 .
\]
Multiplying both sides by $c\,R_2(c,\alpha_1,\alpha_2)$, which is nonzero on $\mathcal{O}$, gives
\begin{equation}\label{eq:6}
  \alpha_1\alpha_2\, R_2(c,\alpha_1,\alpha_2) \;=\; c\,\bigl(4R_1(c,\alpha_1,\alpha_2) - R_2(c,\alpha_1,\alpha_2)\bigr)
  \qquad \text{for all } (c,\alpha_1,\alpha_2) \in \mathcal{O}.
\end{equation}
Both sides of~\eqref{eq:6} are polynomials. Two polynomials that agree on a nonempty open set are identical, so~\eqref{eq:6} holds for all $(c,\alpha_1,\alpha_2) \in \mathbb{R}^3$. Setting $c = 0$ gives
\[
  \alpha_1\alpha_2\, R_2(0,\alpha_1,\alpha_2) = 0 \qquad \text{for all } \alpha_1, \alpha_2 \in \mathbb{R}.
\]
Since this holds for all $(\alpha_1,\alpha_2)$ with $\alpha_1\alpha_2 \neq 0$, the polynomial $R_2(0,\alpha_1,\alpha_2)$ vanishes on a nonempty open set and is therefore the zero polynomial. In particular $R_2(0,0,0) = 0$, contradicting $R_2(0,0,0) \neq 0$.
\end{proof}

\begin{corollary}
    $Do$-calculus is not complete for uID problem.
\end{corollary}
\begin{proof}
Suppose, for contradiction, that do-calculus is complete for uID, and consider $\mathbf{Prob}_{\mathbf{uID}}$ with binary $X, Z, W, T$ as in Theorem~\ref{th:main: prob-uid not do-identifiable}. By Theorem~\ref{th:main:identifiable}, $P(W,T)$ is uID from $\mathcal{Q} = (\mathcal{G}, \{P(X,Z), P(X,T), P(Z,W)\})$. By completeness, there is then a finite sequence of do-calculus rules and standard probability manipulations that transforms $P(w,t)$ into an expression involving only the distributions in $\mathcal{Q}$.

Such a derivation depends only on $\mathcal{G}$ and $\mathcal{Q}$, not on the particular model $\mathcal{M} \in \mathbb{M}(\mathcal{G})$. Moreover, each step uses only addition, subtraction, multiplication and division, where marginalization is a finite sum because the observed variables are binary. Hence, for each $(w,t)$, the resulting expression is a single rational function $\mathcal{F}_{w,t}$ of the entries of $P(X,Z)$, $P(X,T)$ and $P(Z,W)$. Under positivity, this single rational function should be well-defined and equal to $P^{\mathcal{M}}(W=w,T=t)$ for every $\mathcal{M} \in \mathbb{M}(\mathcal{G})$. This contradicts Theorem~\ref{th:main: prob-uid not do-identifiable}.
   
\end{proof}

\section{Conclusion}
We gave a formal, standalone definition of completeness of the do-calculus for an identifiability problem, one that does not require first finding a sound and complete identification algorithm for that problem, and showed that it agrees with the existing completeness results for ID, gID and c-gID. Our main result is that, for the uID problem, a natural extension of gID, the do-calculus is no longer complete: there are causal effects that are uniquely identifiable from the available distributions but cannot be obtained by any sequence of do-calculus rules and probability manipulations. This shows a fundamental limitation of the do-calculus in more general identification settings, and suggests that closing the remaining gaps in the literature will require identification tools beyond these rule-based symbolic derivations.

% References
\bibliography{ref}

%%%%%%%%%%%%%%%%%%%%%%%%%%%%%%%%%%%%%%%%%%%%%%%%%%%%%%%%%%%%

\newpage
\appendix

\section{Appendix}
\subsection{Proofs of section \ref{sec:alignment with previous works}}
\label{app:proofs sec 3}
Let us first provide a formal definition of c-component, which  is essential for understanding Lemmas \ref{lemma:Q operation 1} and \ref{lemma:Q operation 2}.
\begin{definition}[c-component]
     Two variables $X_1, X_2 \in \N$ are said to be connected via a bidirected edge in a semi-Markovian causal graph $\G$ if there exists variable $U \in \U$ such that $(U, X_1), (U, X_2) \in \mathcal{E}$. Then a nonempty set $\X\subseteq\N$ forms a single confounded component (c-component) in $\G_{\X}$ if its vertices are connected by bidirected paths. The c-components of a graph are the maximal sets with this property.
\end{definition}

Another important notion  is the function $Q^{\M}[\S](\n)$. For a given $\S \subseteq \N$ and $\n \in \dom{\N}{}$:
\begin{equation}
    \label{eq:Q function}
    Q^{\M}[\S](\n) := P\left(\n_{\S}|do\left(\n_{\N\setminus\S}\right)\right).
\end{equation}
% and which can be computed as
% \begin{equation*}
%     % \label{eq:Q computation}
%     Q^{\M}[\S](\n) = \sum_{u\in \dom{U}{}}\prod_{S \in \S} P^{\M}(\n_{S}|\n_{\Pa{\S}{}}) \prod_{U \in \U} P^{\M}(\u_{U}).
% \end{equation*}

%These functions represent crucial quantities used by algorithms developed for identifiability problems. 
The following lemmas define two key manipulations of $Q[\cdot](\cdot)$, which form the backbone of the computational steps in the existing algorithms in the identification literature.
\begin{lemma}[\citep{qlemmas}]
    \label{lemma:Q operation 1}
    Let $\W \subseteq \S \subseteq \N$ and $\W' = \S\setminus \W$. If $\W$ is an ancestral set in $\G_{\S}$, then for any $\n \in \dom{N}{}$: 
    % $\sum_{\w' \in \dom{\W'}{}} Q[\S](\n_{\N\setminus \W'}, \w') = Q[\W][\n]$.
    \begin{equation}
        \label{eq:lemma 1}
        \sum_{\w' \in \dom{\W'}{}} Q[\S](\n_{\N\setminus \W'}, \w') = Q[\W](\n)
    \end{equation}
\end{lemma}

\begin{lemma}[\citep{qlemmas}]
    \label{lemma:Q operation 2}
    Let $\H \subseteq \N$, and let $\H_1, \dots, \H_r$ be c-components in the induced graph $\G_{\H}$. Then for any $\n \in \dom{\N}{}$, \\
    1. $Q[\H](\n) = \prod_{i}Q[\H_i](\n)$;\\
    2. Each $Q[\H_i](\n)$ is computable from $Q[\H](\cdot)$.
\end{lemma}
Further, we establish the connection between these Lemmas and $do$-inference rules akin to  \citep{huang2008completeness}.
\begin{restatable}{lemma}{lemmaDoInferenceForQOne}
    \label{lemma:do-inference for Q1}
    Under the assumptions in Lemma \ref{lemma:Q operation 1}, there exists a sequence of $do$-inference rules that transform $Q[\W](\n)$ into $\sum_{\w' \in \dom{\W'}{}} Q[\S](\n_{\N\setminus \W'}, \w')$ and vice versa.
\end{restatable}
\begin{proof}
    We prove the Lemma by explicitly showing that the right-hand side of  \eqref{eq:lemma 1} can be transformed into the left-hand side by a sequence of steps, where every step is either a rule of \textit{do}-calculus or a standard probability manipulation defined in Section \ref{sec: def do-inference}.

    Akin to the proof of Lemma 5 in~\citep{valtortaa06completeness}, we start by assuming that $\mathbf{S} = \mathbf{W}\cup \{X\}$ and $\mathbf{W}$ is an ancestral set in $\mathcal{G}_{\mathbf{S}}$; and then generalize it to an arbitrary $\mathbf{S}$ by applying the lemma iteratively to vertices of $\S\setminus \W$ in the topological order in $\mathcal{G}_{\S}$. Denote the complement of $\mathbf{W}\cup \{X\}$ by $\mathbf{Y} = \mathbf{N} \setminus (\mathbf{W}\cup \{X\})$. We prove that the following identity can be achieved for any realization $\mathbf{n}\in \mathfrak{X}(\mathbf{N})$ with do-inference rules 
    \[
    \sum_{x\in \mathfrak{X}(X)} P (\mathbf{n}_{\mathcal{\mathbf{W}}}, x| do(\mathbf{n}_\mathbf{Y})) = P (\mathbf{n}_{\mathbf{W}}| do(\mathbf{n}_{\mathbf{Y}}, \mathbf{n}_{X})). 
    \]
    Following~\citep{valtortaa06completeness} we have,
    \begin{equation}\label{eq_lemma5_Q}
    \begin{split}
         \sum_{x\in \mathfrak{X}(X)} P (\mathbf{n}_{\mathcal{\mathbf{W}}}, x| do(\mathbf{n}_\mathbf{Y})) &\overset{\text{PR3}}{=} \sum_{x\in \mathfrak{X}(X)} P ( x| \mathbf{n}_{\mathcal{\mathbf{W}}}, do(\mathbf{n}_\mathbf{Y})) P ( \mathbf{n}_{\mathcal{\mathbf{W}}}| do(\mathbf{n}_\mathbf{Y})) \\
        & =\left(\sum_{x\in \mathfrak{X}(X)} P ( x| \mathbf{n}_{\mathcal{\mathbf{W}}}, do(\mathbf{n}_\mathbf{Y}))\right) P ( \mathbf{n}_{\mathcal{\mathbf{W}}}| do(\mathbf{n}_\mathbf{Y})) \\
        & \overset{\text{PR2}}{=} P ( \mathbf{n}_{\mathcal{\mathbf{W}}}| do(\mathbf{n}_\mathbf{Y})) \overset{\text{DR3}}{=} P(\mathbf{n}_{\mathbf{W}}|do(\mathbf{n}_{\mathbf{Y}}, \mathbf{n}_X)).
    \end{split}
    \end{equation}
    %Here we refer to PR2, PR3, and DR3 as to the standard probability manipulation rules and to \textit{do}-calculus rules defined in Section \ref{sec: def do-inference}. 
    Note that the equality in~(\ref{eq_lemma5_Q}) is a standard arithmetic operation. The last transformation, i.e. DR3, is possible  due to the proof presented in~\citep{valtortaa06completeness}. This  presents an explicit transformation of $\sum_{\w' \in \dom{\W'}{}} Q[\S](\n_{\N\setminus \W'}, \w')$ into $Q[\W](\n)$ by application of $do$-inference rules. Since each transformation has an inverse, this concludes the proof of Lemma~\ref{lemma:do-inference for Q1}.
\end{proof}

\begin{restatable}{lemma}{lemmaDoInferenceForQTwo}
    \label{lemma:do-inference for Q2}
    Under the assumptions of Lemma \ref{lemma:Q operation 2}, there is a sequence of $do$-inference rules which transform:\\ \textbf{1.} $Q[\H](\n)$ into $\prod_{i}Q[\H_i](\n)$ and vice versa;\\ \textbf{2.} $Q[\H_i](\n)$ into the expression depending only on $Q[\H](\cdot)$.
\end{restatable}
\begin{proof}
    Similar to the proof of Lemma \ref{lemma:do-inference for Q1}, we will prove both statements \textbf{1} and \textbf{2} by presenting an explicit sequence of $do$-inference rules for each of the statements \textbf{1} and \textbf{2} in the lemma specifically. We will follow the proof steps of Lemma 6~\citep{valtortaa06completeness}.
    
    \textit{Proof of statement} \textbf{1} Similar to the proof in~\citep{valtortaa06completeness}, we will prove statement \textbf{1} by induction on $k$ (size of set $\mathbf{H}$), which is increased by adding a new variable $X$ to the set $\mathbf{H}$ according to the topological order. Define set $\mathbf{E}:= \mathbf{H}\cup \{X\}$, where $X$ has the highest topological order in $\mathcal{G}_{\mathbf{E}}$. Additionally, let $\{\mathbf{H}_i\}_{i=1}^n$denote the c-components in graph $\mathcal{G}_{\mathbf{H}}$ and index $m$ such that $\mathbf{H}_1, \dots, \mathbf{H}_m$ together with $X$ form a single c-component in $\mathcal{G}_\mathbf{E}$ (if $X$ forms a c-component by itself, then $m=0$). Let $\mathbf{H}^i_j$ denote the  set of variables comprised of $\mathbf{H}_j, \mathbf{H}_{j+1}, \dots, \mathbf{H}_i$. We denote the complement of $\mathbf{E}$ by $\mathbf{Y}$, i.e., $\mathbf{Y} = \mathbf{N}\setminus \mathbf{E}$.
    
    \textit{Base of induction}. $\mathbf{E}$ consists of just a single vertex, hence  statement \textbf{1} holds trivially. 
    
    \textit{Induction step}. By the induction hypothesis, we assume that there exists a sequence $\mathbb{S}$ of $do$-inference rules that transform $Q[\mathbf{H}]$ into $\prod_{i}Q[\H_i](\n)$ whenever the size of the set $\mathbf{H}$ is less than $k+1$. Consequently, we will prove the statement of induction for the new set $\mathbf{E}:= \{X\}\cup \mathbf{H}$. More specifically, we have to prove that for any $\mathbf{n}\in \mathfrak{X}(\mathbf{N})$ the following equation holds,
    \[
    \begin{aligned}
    P(\mathbf{n}_{\mathbf{H}}, \mathbf{n}_{X}|do (\mathbf{n}_{\mathbf{Y}})) &= P (\mathbf{n}_{\mathbf{H}^m_1}, \mathbf{n}_X | do (\mathbf{n}_{\mathbf{H}^n_{m+1}},\mathbf{n}_{\mathbf{Y}})) \\
    &\quad\times \prod_{j=m+1}^n P (\mathbf{n}_{\mathbf{H_j}}| do (\mathbf{n}_{\mathbf{H}_1^{j-1}},\mathbf{n}_{\mathbf{H}_{j+1}^n},\mathbf{n}_{\mathbf{Y}}, \mathbf{n}_{X})).
    \end{aligned}
    \]

    From the proof of Lemma 6~\citep{valtortaa06completeness} we have
    \begin{equation}\label{eq_lemma3_pf_lm2_begin}
    \begin{split}
        P(\mathbf{n}_{\mathbf{H}}, \mathbf{n}_{X}|do (\mathbf{n}_{\mathbf{Y}})) 
        \overset{\text{PR3}}{=}&
        P( \mathbf{n}_{X}|\mathbf{n}_{\mathbf{H}}, do ( \mathbf{n}_{\mathbf{Y}})) P(\mathbf{n}_{\mathbf{H}}|do (\mathbf{n}_{\mathbf{Y}}))\\
        \overset{\text{DR3}}{=}
         & P( \mathbf{n}_{X}|\mathbf{n}_{\mathbf{H}}, do ( \mathbf{n}_{\mathbf{Y}})) P(\mathbf{n}_{\mathbf{H}}|do (\mathbf{n}_{X}, \mathbf{n}_{\mathbf{Y}})), 
    \end{split}
    \end{equation}
    where the last transformation DR3 holds according to the Equation (30) in~\citep{valtortaa06completeness}. Note that we can apply the assumption of induction to the right-hand side of the last equation in~(\ref{eq_lemma3_pf_lm2_begin}). This implies,
    \begin{equation*}
    \begin{split}
        \overset{\text{Ind}}{=} 
         P( \mathbf{n}_{X}|\mathbf{n}_{\mathbf{H}}, do ( \mathbf{n}_{\mathbf{Y}})) &\times \prod_{j=1}^m P(\mathbf{n}_{\mathbf{H}_j}|do (\mathbf{n}_{\mathbf{H}_1^{j-1}}, \mathbf{n}_{\mathbf{H}_{j+1}^n}, \mathbf{n}_{\mathbf{Y}}, \mathbf{n}_{X}))\\
         & \times \prod_{j=m+1}^n P(\mathbf{n}_{\mathbf{H}_j}|do (\mathbf{n}_{\mathbf{H}_1^{j-1}}, \mathbf{n}_{\mathbf{H}_{j+1}^n}, \mathbf{n}_{\mathbf{Y}}, \mathbf{n}_{X})).
    \end{split}
    \end{equation*}
    Now we can apply once again the induction step but this time to the middle term 
    \begin{equation*}
    \begin{aligned}
        &\overset{\text{Ind}}{=}
         P( \mathbf{n}_{X}|\mathbf{n}_{\mathbf{H}}, do ( \mathbf{n}_{\mathbf{Y}})) P(\mathbf{n}_{\mathbf{H}_1^m}|do (\mathbf{n}_{\mathbf{H}_{m+1}^n}, \mathbf{n}_{\mathbf{Y}}, \mathbf{n}_{X}))\\
        &\quad\times \prod_{j=m+1}^n P(\mathbf{n}_{\mathbf{H}_j}|do (\mathbf{n}_{\mathbf{H}_1^{j-1}}, \mathbf{n}_{\mathbf{H}_{j+1}^n}, \mathbf{n}_{\mathbf{Y}}, \mathbf{n}_{X})).
    \end{aligned}
    \end{equation*}
    From Equation (33) in~\cite{valtortaa06completeness}, we can delete the action in the middle term, i.e., apply to it the third rule of \textit{do}-calculus
    \begin{equation*}
    \begin{aligned}
        &\overset{\text{DR3}}{=}
         P( \mathbf{n}_{X}|\mathbf{n}_{\mathbf{H}}, do ( \mathbf{n}_{\mathbf{Y}})) P(\mathbf{n}_{\mathbf{H}_1^m}|do (\mathbf{n}_{\mathbf{H}_{m+1}^n}, \mathbf{n}_{\mathbf{Y}}))\\
        &\quad\times \prod_{j=m+1}^n P(\mathbf{n}_{\mathbf{H}_j}|do (\mathbf{n}_{\mathbf{H}_1^{j-1}}, \mathbf{n}_{\mathbf{H}_{j+1}^n}, \mathbf{n}_{\mathbf{Y}}, \mathbf{n}_{X})).
    \end{aligned}
    \end{equation*}
    By Equation (34) in~\citep{valtortaa06completeness}, we can apply the second rule of do-calculus to the left hand side term and get
    \begin{equation*}
    \begin{aligned}
        &\overset{\text{DR2}}{=} 
        P(\mathbf{n}_{X}|\mathbf{n}_{\mathbf{H_1^m}}, do ( \mathbf{n}_{\mathbf{H_{m+1}^n}},  \mathbf{n}_{\mathbf{Y}})) P(\mathbf{n}_{\mathbf{H}_1^m}|do (\mathbf{n}_{\mathbf{H}_{m+1}^n}, \mathbf{n}_{\mathbf{Y}}))\\
        &\quad\times \prod_{j=m+1}^n P(\mathbf{n}_{\mathbf{H}_j}|do (\mathbf{n}_{\mathbf{H}_1^{j-1}}, \mathbf{n}_{\mathbf{H}_{j+1}^n}, \mathbf{n}_{\mathbf{Y}}, \mathbf{n}_{X})).
    \end{aligned}
    \end{equation*}
    And we finish up by applying the chain rule 
    \begin{equation*}
        \overset{\text{PR3}}{=}
        P(\mathbf{n}_{\mathbf{H_1^m}},\mathbf{n}_{X}| do ( \mathbf{n}_{\mathbf{H_{m+1}^n}},  \mathbf{n}_{\mathbf{Y}})) \times \prod_{j=m+1}^n P(\mathbf{n}_{\mathbf{H}_j}|do (\mathbf{n}_{\mathbf{H}_1^{j-1}}, \mathbf{n}_{\mathbf{H}_{j+1}^n}, \mathbf{n}_{\mathbf{Y}}, \mathbf{n}_{X})).
    \end{equation*}
    These transformations form a sequence of $do$-inference rules that transforms $Q[\H](\n)$ into $\prod_{i}Q[\H_i](\n)$. Since each transformation has an inverse, this concludes the proof of the induction step and the proof of the statement \textbf{1} of the lemma.

    \textit{Proof of statement} \textbf{2}. To prove statement \textbf{2}, we will utilize the proof of Lemma 6~\citep{valtortaa06completeness} and show that there exists a sequence of $do$-inference rules that transforms $Q[\H_{i}](\n)$ into the expression that depends only on $Q[\H](\cdot)$.  Let $\mathbf{H}$ consist of vertices $\{V_{h_1}, \dots , V_{h_k}\}$ ordered in the topological order in $\mathcal{G}_{\mathbf{H}}$, i.e., if there is an edge from $V_{h_i}$ to $V_{h_j}$ then $i < j$, and let $\mathbf{H}^{(j)}$ be a variable set $\{V_{h_1}, \dots, V_{h_j}\}$. Denote by $\mathbf{Y}=\mathbf{N}\setminus \mathbf{H}$, the complement of $\mathbf{H}$. 
    \cite{tian2003ID} proved that for any realization $\mathbf{n}\in \mathfrak{X}(\mathbf{N})$
    \begin{equation}\label{eq_lemma_3_Q[H_j]}
        Q[\mathbf{H}_i](\mathbf{n}) = \prod_{\{j| V_{h_j}\in H_i\}} \frac{Q[\mathbf{H}^{(j)}](\mathbf{n})}{Q[\mathbf{H}^{(j-1)}](\mathbf{n})}, 
    \end{equation}
     and
    \begin{equation}\label{eq_lemma_3_marg}
        Q[\mathbf{H}^{(j)}](\mathbf{n}) = \sum_{\mathbf{h} \in \mathfrak{X}(\H \setminus \H^{(j)})} Q[\mathbf{H}](\mathbf{n}_{\mathbf{N}\setminus(\H\setminus\H^{(j)})}, \mathbf{h}). 
    \end{equation}
   Consider Equation~(\ref{eq_lemma_3_marg}), since the vertices in $\mathbf{H}$  are ordered in the topological order, the set
    \[
    \mathbf{H}^{(j)} = \{V_{h_1}, \dots , V_{h_j}\}
    \]
    is ancestral in $\mathcal{G}_{\mathbf{H}}$. Hence, applying Lemma~\ref{lemma:do-inference for Q1} to Equation~(\ref{eq_lemma_3_marg}), implies that there exists a sequence of $do$-calculus rules that transform one side of the equation into the other. Let us now show that the same holds for the equation~(\ref{eq_lemma_3_Q[H_j]}). By the definitions, we have  
    \begin{equation}\label{eq_lemma3_def}
        Q [\mathbf{H}_i](\mathbf{n}) = P (\mathbf{n}_{\mathbf{H}_i} | do(\mathbf{n}_{\mathbf{H}\setminus \mathbf{H}_i}, \mathbf{n}_{\mathbf{Y}})).
    \end{equation}
    Let $i_1< \dots < i_l$ denote the set of indices of the variables in $\mathbf{H}_i$, i.e., $\mathbf{H}_i = \{V_{h_{i_1}}, \dots, V_{h_{i_l}}\}$. Denote by $\mathbf{V}_j^m$ the variable set consisting of $V_{h_{i_{j}}}, V_{h_{i_{j+1}}}, \dots, V_{h_{i_{m}}}$. From the chain rule Equation~(\ref{eq_lemma3_def}) can be rewritten as 
    \begin{equation}\label{eq_lemma_3_prod}
        P (\mathbf{n}_{\mathbf{H}_i} | do(\mathbf{n}_{\mathbf{H}\setminus \mathbf{H}_i}, \mathbf{n}_{\mathbf{Y}})) \overset{\text{PR3}}{=} P (v_{h_{i_l}}| \mathbf{n}_{\mathbf{V}_1^{l-1}}, do(\mathbf{n}_{\mathbf{H}\setminus \mathbf{H}_i}, \mathbf{n}_{\mathbf{Y}})) P ( \mathbf{n}_{\mathbf{V}_1^{l-1}} | do(\mathbf{n}_{\mathbf{H}\setminus \mathbf{H}_i}, \mathbf{n}_{\mathbf{Y}})), 
    \end{equation}
    where $v_{h_{j}}$ is the realization of $V_{h_j}$ in $\mathbf{n}$. We can continue applying the chain rule to the right hand side term, and we get the following product 
    \begin{equation*}
        \overset{\text{PR3}}{=}\prod_{j=1}^l P (v_{h_{i_j}}| \mathbf{n}_{\mathbf{V}_1^{j-1}}, do(\mathbf{n}_{\mathbf{H}\setminus \mathbf{H}_i}, \mathbf{n}_{\mathbf{Y}})), 
    \end{equation*}
    where $\mathbf{V}_1^0$ is an empty set. Now let us show that the following identity holds for any $j$
    \begin{equation}\label{eq_lemma_3_P(v|...)}
        P (v_{h_{i_j}}| \mathbf{n}_{\mathbf{V}_1^{j-1}}, do(\mathbf{n}_{\mathbf{H}\setminus \mathbf{H}_i}, \mathbf{n}_{\mathbf{Y}})) = P (v_{h_{i_j}}| \mathbf{n}_{\mathbf{H}^{(i_j-1)}}, do( \mathbf{n}_{Y})). 
    \end{equation}
    We can split up $\mathbf{H}\setminus \mathbf{H}_i$ into two variable sets $\mathbf{W}^+$ and $\mathbf{W}^-$, such that $\mathbf{W}^+$ consists of the variables of $\H\setminus \H_i$ with higher topological orders in $\mathcal{G}_{\mathbf{H}}$ than $V_{h_{i_j}}$ and, respectively, $\mathbf{W}^-$ consists of the variables of $\H\setminus\H_i$ with lower topological orders. Note that $\mathbf{W}^+(\mathbf{V}_1^{j-1}) = \mathbf{W}^+$ in $\G_{\overline{\mathbf{W}^-},\overline{\mathbf{Y}}}$\footnote{Where $\mathbf{W}^+(\mathbf{V}_1^{j-1})$ is defined similar to $\Z(\W)$ in the definition of DR3.}, that is there are no nodes in $\W^+$ that are ancestors of any node in $\V^{j-1}_1$ in $\G_{\overline{\mathbf{W}^-},\overline{\mathbf{Y}}}$.  Indeed, since all incoming edges of $\mathbf{Y}$ are deleted in $\G_{\overline{\mathbf{W}^-},\overline{\mathbf{Y}}}$ it means that there exists a node in $\mathbf{W}^+$, which is an ancestor of a node in $\mathbf{V}_1^{j-1}$ in $\G_{\mathbf{H}}$. This  contradicts the topological order of vertices in $\H$. 
    
    We can show that the following independence holds:
    \begin{equation}\label{eq_lemma_3_ind_1}
        V_{h_{i_j}}\independent \mathbf{W}^+ | \mathbf{W}^-, \mathbf{Y}, \mathbf{V}_1^{j-1}~~\text{in}~~~\G_{\overline{\mathbf{W}^-},\overline{\mathbf{Y}}, \overline{\mathbf{W}^+}}.
    \end{equation}
    This is because if there is a d-connected path from $V_{h_{i_j}}$ to a node $Z$ in $\mathbf{W}^+$ in $\G_{\overline{\mathbf{W}^-},\overline{\mathbf{Y}}, \overline{\mathbf{W}^+(\mathbf{V}_1^{j-1})}}$, then it cannot pass through any node of  $\mathbf{W}^-$ or $\mathbf{Y}$. Let $Z'$ be adjacent on this path to $Z$. It is observable, since there are no incoming edges to both $\mathbf{U}$ and $\mathbf{W}^{+}$ in $\mathcal{G}_{\overline{\mathbf{W}^+}}$. Note that $Z'$ cannot be in $\mathbf{W}^+$ as there are no edges between the vertices in $\G_{\overline{\mathbf{W}^+}}$.  It  cannot be in $V_{h_{i_j}}$ either since it violates the topological order. Therefore, $Z'\in \mathbf{V}_{j+1}^l$ and the edge is oriented from $Z$ to $Z'$. On the other hand, the topological order requires that the first edge of the path goes out from $V_{h_{i_j}}$. Therefore, the path contains a collider given that a middle vertex is  in $\mathbf{V_{j+1}^l}$. But we have
    \[
    (\mathbf{W}^-, \mathbf{Y}, \mathbf{V}_1^{j-1}) \cap De_{\G_{\overline{\mathbf{W}^-},\overline{\mathbf{Y}}, \overline{\mathbf{W}^+}}} (\mathbf{V}_{j+1}^l) = \emptyset,  
    \]
    which means that the path we considered from $V_{h_{i_j}}$ to $Z$ remains d-separated. The independence in~(\ref{eq_lemma_3_ind_1}), allows us to apply the third rule of \textit{do}-calculus to the left hand side of~(\ref{eq_lemma_3_P(v|...)}) and get 
    \begin{equation}\label{eq_lemma_3_DR3}
        P (v_{h_{i_j}}| \mathbf{n}_{\mathbf{V}_1^{j-1}}, do(\mathbf{n}_{\mathbf{H}\setminus \mathbf{H}_i}, \mathbf{n}_{\mathbf{Y}})) \overset{\text{DR3}}{=} P (v_{h_{i_j}}| \mathbf{n}_{\mathbf{V}_1^{j-1}}, do(\mathbf{n}_{\mathbf{W}^-}, \mathbf{n}_{Y})).
    \end{equation}

    Just as before, we also have the following d-separation 
    \begin{equation}\label{eq_lemma_3_ind_2}
        V_{h_{i_j}}\independent \mathbf{W}^- | \mathbf{Y}, \mathbf{V}_1^{j-1}~~\text{in}~~~\G_{\overline{\mathbf{Y}}, \underline{\mathbf{W}^-}}.
    \end{equation}
    Indeed, let us assume a contradiction, namely that there is a d-connected path from $Z\in \mathbf{W}^-$ to $\mathbf{V}_{h_{i_j}}$ in $\G_{\overline{\mathbf{Y}}, \underline{\mathbf{W}^-}}$, then it clearly cannot pass through $\mathbf{Y}$. We can also assume that this path has the shortest length. Let $Z'$ be the nearest node to $Z$ on that path, note that we have an edge going from $Z'$ to $Z$ as there are no outgoing edges in $\G_{\underline{\mathbf{W}^-}}$. The node $Z'$ cannot be in $\mathbf{W}^-$ as there are no edges in between vertices in $\G_{\underline{\mathbf{W^-}}}$ and it cannot be $V_{h_{i_j}}$ as it violates the topological order. So, we get that either $Z'\in \mathbf{V}_{j+1}^l$ or $Z'\in \mathbf{U}$. Firstly, we assume that $Z'$ is a latent variable. Since $\mathbf{H}_i$ is a $c$-component, any node adjacent to $Z'$ in $\G_{\overline{\mathbf{Y}}, \underline{\mathbf{W}^-}}$ lies either in $\mathbf{W}^{+}$ or $\mathbf{W}^{-}$. If the adjacent node to $Z'$ on the path, denote it by $Z''$, lies in $\mathbf{W}^{-}$, then it contradicts the assumption that the path is of minimal length. Thus, we obtain either $Z'\in \mathbf{V}_{j+1}^l$ or $Z''\in \mathbf{W}^{+}$. As before, due to the topological order, the path contains a collider which is in $\mathbf{V}_{j+1}^l \cup \mathbf{W}^+$. We have 
    \[
    (\mathbf{Y}, \mathbf{V}_1^{j-1}) \cap De_{\G_{\overline{\mathbf{Y}}, \underline{\mathbf{W}^-}}} (\mathbf{V}_{j+1}^l\cup \mathbf{W}^{+}) = \emptyset,  
    \]
    that concludes the proof of the d-separation given in~(\ref{eq_lemma_3_ind_2}). From the independence~(\ref{eq_lemma_3_ind_2}), it follows that we can apply the second rule of do-calculus to ~(\ref{eq_lemma_3_DR3}) and get that 
    \[
      P (v_{h_{i_j}}| \mathbf{n}_{\mathbf{V}_1^{j-1}}, do(\mathbf{n}_{\mathbf{W}^-},\mathbf{n}_{Y})) \overset{\text{DR2}}{=} P (v_{h_{i_j}}| \mathbf{n}_{\mathbf{V}_1^{j-1}},\mathbf{n}_{\mathbf{W}^-}, do( \mathbf{n}_{Y}))=P (v_{h_{i_j}}| \mathbf{n}_{\mathbf{H}^{(i_j-1)}}, do( \mathbf{n}_{Y})), 
    \]
    where the last equality follows from the definition of the variable set $\mathbf{W}^{-}$. This proves the statement in (\ref{eq_lemma_3_P(v|...)}).

    Finally, we show that the following identity holds for any $r$, in particular when $r$ is of the form $i_j$
    \begin{equation}\label{eq_lemma_3_fraction}
        P(\mathbf{n}_{V_{h_r}}| \mathbf{n}_{\mathbf{H}^{(r-1)}}, do(\mathbf{n}_{\mathbf{Y}})) = \frac{Q[\mathbf{H}^{(r)}](\mathbf{n})}{Q[\mathbf{H}^{(r-1)}](\mathbf{n})}. 
    \end{equation}
    From the conditioning rule, we have that 
    \begin{equation*}
        P(\mathbf{n}_{V_{h_r}}| \mathbf{n}_{\mathbf{H}^{(r-1)}}, do(\mathbf{n}_{\mathbf{Y}})) \overset{\text{PR4}}{=} \frac{P(\mathbf{n}_{\mathbf{H}^{(r)}} | do(\mathbf{n}_{\mathbf{Y}}))}{P(\mathbf{n}_{\mathbf{H}^{(r-1)}}| do(\mathbf{n}_{\mathbf{Y}}))}. 
    \end{equation*}
    Apply the marginalization rule to both numerator and denominator to get
    \begin{equation*}
    \begin{aligned}
        &\overset{\text{PR2}}{=} \frac{\sum_{\mathbf{v}\in \mathfrak{X}(\mathbf{\mathbf{H}\setminus \mathbf{H}^{(r)}})} P(\mathbf{n}_{\mathbf{H}^{(r)}}, \mathbf{v} | do(\mathbf{n}_{\mathbf{Y}}))}{\sum_{(\mathbf{v}, v_r) \in \mathfrak{X}( \mathbf{H}\setminus \mathbf{H}^{(r-1)})} P(\mathbf{n}_{\mathbf{H}^{(r-1)}}, \mathbf{v}, v_r| do(\mathbf{n}_{\mathbf{Y}}))}\\
        &= \frac{\sum_{\mathbf{v}\in \mathfrak{X}(\mathbf{\mathbf{H}\setminus \mathbf{H}^{(r)}})} Q[\mathbf{H}](\mathbf{n}_{\mathbf{H}^{(r)}}, \mathbf{v})}{\sum_{(\mathbf{v}, v_r) \in \mathfrak{X}( \mathbf{H}\setminus \mathbf{H}^{(r-1)})} Q[\mathbf{H}](\mathbf{n}_{\mathbf{H}^{(r-1)}}, \mathbf{v}, v_r)}. 
    \end{aligned}
    \end{equation*}
    Applying (\ref{eq_lemma_3_marg}) to both numerator and denominator results in~(\ref{eq_lemma_3_fraction}).

    Combining everything together, we obtained the following sequence of \textit{do}-inference rules that computes $Q[\mathbf{H}_i](\mathbf{n})$ in terms of $Q[\H](\cdot)$
    \begin{equation*}
    \begin{split}    
        Q[\mathbf{H}_i](\mathbf{n}) \overset{(\ref{eq_lemma_3_prod})}{=} \prod_{j=1}^l P (v_{h_{i_j}}| \mathbf{n}_{\mathbf{V}_1^{j-1}}, do(\mathbf{n}_{\mathbf{H}\setminus \mathbf{H}_i}, \mathbf{n}_{\mathbf{Y}})) &\overset{(\ref{eq_lemma_3_P(v|...)})}{=}\prod_{j=1}^l P (v_{h_{i_j}}| \mathbf{n}_{\mathbf{H}^{(i_j-1)}}, do( \mathbf{n}_{Y}))\\
        &\overset{(\ref{eq_lemma_3_fraction})}{=}\prod_{j=1}^l \frac{Q[\mathbf{H}^{(i_j)}](\mathbf{n})}{Q[\mathbf{H}^{(i_j-1)}](\mathbf{n})}.
    \end{split}
    \end{equation*}
    
\end{proof}

\begin{algorithm}[t]
    \caption{ID\_Single: identification of $Q[\S]$ from $(Q[\A], \G)$}
    \label{alg:compute Q}
    \textbf{Input:} $\G$, $\emptyset\ne\S\subseteq\A$ with $\G_\S$ a single c-component, $Q[\A](\cdot)$
    
    \begin{algorithmic}[1]
    \STATE $\A_0 \gets \A$
    \STATE $\S_0 \gets$ c-component in $\G_{\A_0}$ that contains $\S$ 
    \STATE $k \gets 0$, $\S_{-1}\gets\emptyset$
    \WHILE{$\S_k\neq \S$ and $\S_k\neq\S_{k-1}$}
    \STATE $k \gets k+1$
    \STATE $\A_k \gets \Anc{\S}{\G_{\S_{k-1}}}$
    \STATE $\S_k \gets$ c-component in $\G_{\A_k}$ that contains $\S$
    \ENDWHILE
    \IF{$\S_k = \S_{k-1}$}
    \STATE \textbf{Return:} False
    \ENDIF
    \STATE \textbf{Return:} True
    \end{algorithmic}
\end{algorithm}

Suppose $\A$ is a subset of $\N$ and $\emptyset\ne\S\subseteq\A$ forms a single c-component in $\G_\S$. Next,  we introduce Algorithm ~\ref{alg:compute Q} that determines whether distribution $Q[\S]$ is identifiable from query $\mathcal{Q}=(\G, Q[\A])$. This simple algorithm is developed based on the aforementioned lemmas, where $Q[\A_k]$ can be computed from $Q[\S_{k-1}]$ in line 6 according to Lemma~\ref{lemma:Q operation 1}, and $Q[\S_k]$ can be computed from $Q[\A_k]$ in line 7 according to the  Lemma~\ref{lemma:Q operation 2}. In fact, Algorithm~\ref{alg:compute Q} is a crucial subroutine of existing algorithms in the classical literature on identifiability problems, such as \citep{shpitser2006identification, huang2006identifiability, kivva2022revisiting}.

\begin{restatable}{lemma}{lemmaAlgQ}
    \label{lemma_alg_Q[S]}
    If Algorithm~ \ref{alg:compute Q} returns True, then $Q[\S]$ is do-identifiable from query $\mathcal{Q}=(\G, Q[\A])$.
\end{restatable}
\begin{proof}
    Let us define functions $\mathcal{H}_{1}(\cdot)$, $\mathcal{H}_{2}(\cdot)$ as follows:\\
    % \begin{enumerate}
    %     \item $\mathcal{L}_1(Q[\W][\v], Q[\S])$ is the sequence of $do$-inference rules that transforms right-hand side of Eq. \eqref{eq:lemma 1} into left-hand side;
    %     \item $\L_{2}(Q[\H_i](\v), Q[\H])$ is a sequence of $do$-inference rules that transforms $Q[\H_i](\v)$ into the formula that only depends on $Q[\H](\cdot)$.
    % \end{enumerate}
    1. $\mathcal{H}_1(Q[\W](\n), Q[\S])$ is a function that uses the sequence of $do$-inference rules to transform the right-hand side of the equation given by Lemma~\ref{lemma:Q operation 1} into the left-hand side;\\
    % 2. $\L_{2, 1}(Q[\H](\n))$ is a sequence of $do$-inference rules that transforms $Q[\H](\n)$ into $\prod_{i}Q[\H_i](\n)$;\\
    2. $\mathcal{H}_{2}(Q[\H_i](\n), Q[\H])$ is a function that uses the sequence of $do$-inference rules to transform $Q[\H_i](\n)$ into the formula that only depends on $Q[\H](\cdot)$ given by Lemma~\ref{lemma:do-inference for Q2}.
    
    Suppose $(\A_0, \S_0, \dots, \A_k, \S_k)$ is a sequence of sets generated by Alg \ref{alg:compute Q}. Note that $\S_k=\S$ and from Lemma \ref{lemma:Q operation 2} we have $Q[\S_k](\n) = \mathcal{H}_{2}(Q[\S](\n), Q[\A_k])$. Additionally, note that $\mathcal{H}_{2}(Q[\S_k](\n), Q[\A_k])$ is expressed only by terms of form $Q[\A_k]$. Based on Lemma \ref{lemma:Q operation 1}, each term $Q[\A_k](\cdot)$ can be expressed as $Q[\A_k] = \mathcal{H}_{1}(Q[\A_{k}], Q[\S_{k-1}])$. Similarly, $Q[\A_k]$ can be expressed by the terms $Q[\S_{k-1}]$. Consequently, we may continue this procedure until $Q[\S_0]$ is expressed by the terms $Q[\A_0]$, i.e. $Q[\A]$. Recall that each function $\mathcal{H}_{1}$ and $\mathcal{H}_{2}$ corresponds to some sequence of $do$-inference rules. This implies that we can start with $Q[\S]$ and via composition of functions $\mathcal{H}_{1}$ and $\mathcal{H}_{2}$ (i.e. $do$-inference rules) reach an expression that is written in terms of $Q[\A]$. That concludes the proof.
\end{proof}

It is well-known that Algorithm \ref{alg:compute Q} is sound and complete \citep{huang2006identifiability, kivva2022revisiting}. Hence, $do$-calculus rules are complete for the identifiability problem solved by Algorithm \ref{alg:compute Q}. Furthermore, we demonstrate that this result extends to the entire classical line of research on the identifiability problem.

\begin{definition}[ID \cite{shpitser2006identification}]
    Suppose $\X$ and $\Y$ are disjoint subsets of $\N$.  The causal effect $P(\y|do(\x))$ is said to be identifiable (ID) in $\G$ if for any $\x\in \dom{\X}{}$, $\y \in \dom{\Y}{}$ the causal effect $P^{\M}(\y|do(\x))$ is uniquely computable from $P^{\M}(\N)$ in any causal model $\M \in \mathbb{M}(\G)$.
\end{definition}
\begin{definition}[gID \cite{kivva2022revisiting}]
    Suppose $\X$ and $\Y$ are disjoint subsets of $\N$, $\mathcal{Z} = \{\Z_i\}_{i=0}^{m}$ be a collection of subsets of $\N$ and let $\G$ be a given causal graph corresponding to the set of variables $\N$. The causal effect $P(\y|do(\x))$ is said to be g-identifiable from $\mathcal{Z}$ in $\G$ if $P^{\M}(\y|do(\x))$ is uniquely computable from interventional distributions $\{P^{\M}\left(\Z_{i}|do(\N\setminus \Z_{i})\right)\}_{i=0}^{m}$ in any causal model $\M \in \mathbb{M}(\G)$.
\end{definition}

\begin{definition}[c-gID \cite{kivva2023identifiability}]
    Suppose $\X$, $\W$ and $\Y$ are disjoint subsets of $\N$, let $\mathcal{Z} = \{\Z_i\}_{i=0}^{m}$ be a collection of subsets of $\N$ and let $\G$ be a given causal graph on the set of variables $\N$. The causal effect $P(\y|do(\x), \w)$ is c-gID from $\mathcal{Z}$ in $\G$ if $P^{\M}(\y|do(\x), \w)$ is uniquely computable from interventional distributions $\{P^{\M}\left(\Z_{i}|do(\N\setminus \Z_{i})\right)\}_{i=0}^{m}$ in any causal model $\M \in \mathbb{M}(\G)$.
\end{definition}

% \begin{customtheorem}{\ref{thm_completeness for gID}}
%      $do$-calculus is complete for the ID, gID and c-gID problems studied in \citep{shpitser2006identification, kivva2022revisiting, kivva2023identifiability}.
% \end{customtheorem}
\thDoCalculusWellDefined*
\begin{proof} Let us prove the completeness of \textit{do}-calculus for c-gID problem. Suppose $\X, \Y, \W \subset \N$ are disjoint subsets of variables in $\G$ and $\mathcal{Z} = \{\Z_i\}_{i=0}^m$ is a collection of subsets in $\N$ in a given $\G$. Then in terms of c-gID problem query is defined as $\mathcal{Q} := (\G, \{P\left(\Z_{i}|do(\N\setminus \Z_{i})\right)\}_{i=0}^{m})$.  To show that $do$-calculus is complete for c-gID, we show that when the causal effect $P(\y | do(\x), \w)$ is identifiable from query $\mathcal{Q}$, then it is $do$-identifiable from query $\mathcal{Q}$. In~\cite{kivva2023identifiability} it was proved that Alg.~\ref{alg_cgid}, provided below, is sound and complete for c-gID. 

\begin{algorithm}[H]
    \caption{c-gID of $P(\mathbf{Y}|do(\mathbf{X}), \W)$ from $(\mathcal{Z}, \G)$}
    \label{alg_cgid}
    \textbf{Input:} $\mathbf{X}$, $\mathbf{Y}$, $\W$, $\mathcal{Z} = \{\mathbf{Z}_i\}_{i=0}^m$, $\G$ \\
    \begin{algorithmic}[1]
    \STATE $\mathbf{T} \gets$ \textbf{MaxBI}$(\X, \Y, \W, \G)$
    \STATE \textbf{Return GID}$(\X\cup \T, \Y \cup (\W\setminus \T), \mathcal{Z}, \G)$ 
    \end{algorithmic}
    \begin{algorithmic}[1]
    \STATE \textbf{Function} \textbf{MaxBI}$(\X, \Y, \W, \G)$
    \STATE $\T \gets \emptyset$
    \FOR{$W$ in $\W$}
    \IF{$\Y \independent W \mid \X,\W\setminus\{W\}$ in $\G_{\overline{\X},\underline{W}}$}
    \STATE $\T \gets \T \cup \{W\}$
    \ENDIF
    \ENDFOR
    \STATE \textbf{Return:} $\T$
    \end{algorithmic}
\end{algorithm}

If Alg.~\ref{alg_cgid} returns True, then $P(\y,\w_{\W\setminus\T}\mid do(\x,\w_{\T}))$ is gID from $\mathcal{Z}$ in $\G$, where $\T$ is returned by \textbf{MaxBI}. By the action/observation exchanges defining $\T$, followed by conditioning and marginalization,
\[
P(\y\mid do(\x),\w)
=\frac{P(\y,\w_{\W\setminus\T}\mid do(\x,\w_{\T}))}
{\sum_{\y'\in\dom{\Y}{}}P(\y',\w_{\W\setminus\T}\mid do(\x,\w_{\T}))}.
\]
The denominator is positive under our positivity assumption. Thus, a $do$-inference expression for the joint effect on the right yields one for the conditional target, so it suffices to prove completeness for gID.

In~\cite{kivva2022revisiting} it is proved that the Alg. $\ref{alg_gid}$ that determines whether the causal effect $P(\y | do (\x))$ is identifiable from query $\mathcal{Q}=(\G, \{P\left(\Z_{i}|do(\N\setminus \Z_{i})\right)\}_{i=0}^{m})$ is sound and complete. 

\begin{algorithm}[H]
    \caption{gID of $P(\mathbf{Y}|do(\mathbf{X}))$ from $(\mathcal{Z}, \G)$}
     \label{alg_gid}
    \textbf{Input:} $\mathbf{X}$, $\mathbf{Y}$, $\mathcal{Z} = \{\mathbf{Z}_i\}_{i=0}^m$, $\G$ \\
    \begin{algorithmic}[1]
    \STATE $\mathbf{S} \gets Anc_{\G_{\mathbf{N}\setminus \mathbf{\mathbf{X}}}}(\mathbf{Y})$
    \STATE $\{\mathbf{S}_1, \dots , \mathbf{S}_l\}\gets$ c-components of $\S$ 
    \FOR{$j$ from $1$ to $l$}
    \IF{\textbf{GID$\_$Single}$(\S_j, \mathcal{Z}, \mathcal{G})=$False}
    \STATE \textbf{Return:} False
    \ENDIF
    \ENDFOR
    \STATE \textbf{Return:} True
    \end{algorithmic}
    \begin{algorithmic}[1]
        \STATE \textbf{Function GID$\_$Single}$(\S, \mathcal{Z} = \{\Z_i\}_{i=0}^m, \mathcal{G})$
        \FOR{$i$ from $0$ to $m$}
        \IF{$\S\subseteq \Z_i$ and \textbf{ID$\_$Single}$(\S, \G[\mathbf{Z_i}])=$True}
        \STATE \textbf{Return:} True
        \ENDIF
        \ENDFOR
        \STATE \textbf{Return:} False
    \end{algorithmic}
\end{algorithm}
Note that \textbf{ID$\_$Single}$(\S, \G[\mathbf{Z_i}])$ is a function given by Alg.~\ref{alg:compute Q} which determines the identification of $Q[\S]$ from query $(\G, Q[\Z_i])$. 

Suppose Alg.~\ref{alg_gid} returns True, namely that the causal effect of $P(\Y| do (\X))$ is identifiable from query $\mathcal{Q}$. Alg. \ref{alg_gid} implies that \textbf{GID$\_$Single}$(\S, \mathcal{Z} = \{\Z_i\}_{i=0}^m, \mathcal{G})$ returns True for every c-component of $\S$. Or equivalently, for every $i$, the causal effect $Q[\S_i](\mathbf{n})$ is identifiable from query $(\G, \{ Q[\Z_j]\})$ for some $j$, i.e. Alg \ref{alg:compute Q} returns True. From Lemma~\ref{lemma_alg_Q[S]} it follows that for every $i$ the causal effect $Q[\S_i](\mathbf{n})$ is \textit{do}-identifiable from query $\mathcal{Q}$.

Following~\cite{kivva2022revisiting}, $P(\y| do (\x))$ is derived as follows. Firstly, we apply the marginalization rule
\begin{equation}
    \label{eq: completeness g-id step 1}
    P(\y | do(\x)) \overset{\text{PR2}}{=} \sum_{\n_{\S\setminus \Y}\in \mathfrak{X}(\S\setminus \Y)}\sum_{\mathbf{n'}\in \mathfrak{X}((\mathbf{N}\setminus \X)\setminus \S)} P(\y,\n_{\S\setminus \Y}, \n'  | do(\x)).
\end{equation}
Since $\S$ is an ancestral set in $\G_{\N \setminus \X}$, from Lemma~\ref{lemma:do-inference for Q1}, there exist $do$-inference rules that transform \eqref{eq: completeness g-id step 1} into 
\begin{equation}
\label{eq: completeness g-id step 2}
= \sum_{\n_{\S\setminus \Y}\in \mathfrak{X}(\S\setminus \Y)} P(\y,\n_{\S\setminus \Y}| do(\x, \n')) 
=
  \sum_{\n_{\S\setminus \Y}\in \mathfrak{X}(\S\setminus \Y)} Q[\S](\y, \mathbf{n}_{\S \setminus \Y}, \n', \x), 
\end{equation}
where the last equation is just rewriting the definition. Finally, from Lemma~\ref{lemma:do-inference for Q2}, there exists the sequence of $do$-inference rules with which we obtain
\[
= \sum_{\n_{\S\setminus\Y}\in\dom{\S\setminus\Y}{}}\prod_{i=1}^{l} Q[\S_i] (\y, \mathbf{n}_{\S \setminus \Y}, \n', \x). 
\]
Since for every $i$ the causal effect $Q[\S_i]$ is \textit{do}-identifiable from query $\mathcal{Q}$, consequently $P(\y|do(\x))$ is also $do$-identifiable from query $\mathcal{Q}$. It finishes the proof that $do$-calculus is complete for gID problem and thus $do$-calculus is complete for c-gID problem as well. Since ID is a particular instance of gID this finishes the proof that $do$-calculus is complete for ID, gID, and c-gID problems. 
\end{proof}

\subsection{Proofs of section \ref{sec:identifiability}}
\label{app:proofs sec 4}
\begin{customprp}{\ref{prop:sec4:non-id}}
    In $\mathbf{Prob_{uID}}$ the causal effects $P^{\mathcal{M}}(T| Z)$, $P^{\mathcal{M}}(W| X)$ and $P^{\mathcal{M}}(X, W, Z, T)$ are not identifiable from $\mathcal{Q}$ in terms of Definition~\ref{def:general form}.
\end{customprp}
\begin{proof}
    To prove the proposition, it is enough, for each target, to construct two positive SCMs $\mathcal{M}_1, \mathcal{M}_2$ such that $\mathcal{Q}^{\M_1} = \mathcal{Q}^{\M_2}$ while the target distributions differ. Let $C = \left[P(x_i, z_j)\right]_{i, j}$ and $P^{\mathcal{M}_1}(X, W, Z, T)$ be a positive distribution in $\mathcal{M}_1$ such that $\dim (\ker C^{\mathcal{M}_1})>0$, see Remark~\ref{remark_1} for the existence. Fix two distinct realizations $t_0,t_1$ of $T$.

    In the new model $\mathcal{M}_2$, we retain $P(X,Z)$ and $P(W|X)$ as in $\M_1$. 
    Also for any realization $t'$ of $T$ outside $\{t_0,t_1\}$ we define 
    \[
    P^{\mathcal{M}_2}(t'|z) : = P^{\mathcal{M}_1}(t'|z), 
    \]
    for any realization $z$. 
    We change only the vectors $\mathbf{a}^{\mathcal{M}_1} := [P^{\mathcal{M}_1}(t_0|z_j)]_j$ and $\mathbf{b}^{\mathcal{M}_1} := [P^{\mathcal{M}_1}(t_1|z_j)]_j$. By the factorization induced by the graph, we obtain
    \begin{equation}\label{eq_P(x,y,z,t)}
        P(X,W,Z,T) = P(X,Z)P(W|X) P(T|Z). 
    \end{equation}
    Therefore, positive perturbed vectors whose sum is unchanged give a well-defined conditional distribution and a positive joint distribution for the model $\mathcal{M}_2$. For this we introduce the following claim.

    \begin{claim}\label{claim_1}
        Let $C\colon \mathbb{R}^m \to \mathbb{R}^n$ be a linear operator such that $\dim (\ker C)>0$, and let $\mathbf{a},\mathbf{b}\in\mathbb{R}^m$ have positive entries. Then there exist positive vectors $\hat{\mathbf{a}}\neq\mathbf{a}$ and $\hat{\mathbf{b}}\neq\mathbf{b}$ satisfying
    \begin{equation}\label{eq_claim_1}
        C\hat{\mathbf{a}}=C\mathbf{a},\qquad C\hat{\mathbf{b}}=C\mathbf{b},\qquad \hat{\mathbf{a}}+\hat{\mathbf{b}}=\mathbf{a}+\mathbf{b}.
    \end{equation}
    \end{claim}

    Apply Claim~\ref{claim_1} to $C^{\mathcal{M}_1},\mathbf{a}^{\mathcal{M}_1},\mathbf{b}^{\mathcal{M}_1}$ and set $\mathbf{a}^{\mathcal{M}_2}=\hat{\mathbf{a}}$ and $\mathbf{b}^{\mathcal{M}_2}=\hat{\mathbf{b}}$. Equation~(\ref{eq_claim_1}) preserves positivity and normalization of $P(T|Z)$ and ensures that $P^{\mathcal{M}_1}(X,T)=P^{\mathcal{M}_2}(X,T)$. Since $P(X,Z)$ and $P(W|X)$ are unchanged, $P(Z,W)$ is also unchanged, so $\mathcal{Q}^{\mathcal{M}_1}=\mathcal{Q}^{\mathcal{M}_2}$. At the same time, $P^{\mathcal{M}_1}(t_0|z')\neq P^{\mathcal{M}_2}(t_0|z')$
    for some realization $z'$, which also implies $P^{\mathcal{M}_1}(X,W,Z,T)\neq P^{\mathcal{M}_2}(X,W,Z,T)$ based on equation~(\ref{eq_P(x,y,z,t)}). This finishes the proof that $P^{\mathcal{M}}(T|Z)$ is not identifiable from the observed distributions. Symmetrically, using $C^T$ and two outcomes of $W$, a separate pair of models shows that $P^{\mathcal{M}}(W|X)$ is not identifiable; for this we start with a distribution satisfying $\dim (\ker (C^{\mathcal{M}_1})^{T})>0$.
\end{proof}
\begin{remark}\label{remark_1}
    Let us construct an example of possible matrix $C$ such that $c_{ij}>0$, $\sum c_{ij} = 1$ and $\dim (\ker C)>0$ (or $\dim (\ker (C)^{T})>0$). In other words, $C$ can be viewed as a linear operator $C\colon \mathbb{R}^m \to \mathbb{R}^n$ for some integer numbers $m > 1,  n > 1$. Let $\mathbf{v}\in \mathbb{R}^m$ with all positive entries then we define matrix $C$ as follows 
    \[
    C^T = \begin{pmatrix}
        \lambda_1 \mathbf{v} & \lambda_2 \mathbf{v} & \dots & \lambda_n \mathbf{v}
    \end{pmatrix}, 
    \]
    where $\lambda_i\in \mathbb{R}$ are positive scalars chosen in a way that $\sum c_{ij} = 1$. Thus, we obtain that 
    \[
    \dim(\ker(C)) = m - \dim(\text{im}(C)) = m-1 > 0,\qquad \dim(\ker(C^T))=n-1>0. 
    \]
    This gives a family of matrices that satisfies the condition $\dim(\ker C) > 0$, $c_{ij}>0$ and $\sum c_{ij} = 1$. Symmetrically, we obtain the family of matrices for the condition $\dim(\ker(C^T)) > 0$. Every such $C$ is realizable by a graph-compatible SCM: let the latent variable $U=(X,Z)$ have distribution $C$, let $X,Z$ be its coordinate projections, and generate $W,T$ from any positive conditional distributions $P(W|X),P(T|Z)$ using independent noise. The resulting joint distribution is positive and satisfies equation~(\ref{eq_P(x,y,z,t)}).
\end{remark}
\begin{proof}[Proof of Claim~\ref{claim_1}]
    Choose a non-zero vector $\tilde{\mathbf{v}}\in\ker C$ and a scalar $\lambda$ such that
    \[
    0<|\lambda|<\min_{j:\tilde{v}_j\neq0}\frac{\min(a_j,b_j)}{|\tilde{v}_j|}.
    \]
    Define $\hat{\mathbf{a}}=\mathbf{a}+\lambda\tilde{\mathbf{v}}$ and $\hat{\mathbf{b}}=\mathbf{b}-\lambda\tilde{\mathbf{v}}$. Both vectors are positive and differ from the original vectors. Since $C\tilde{\mathbf{v}}=0$, they satisfy equation~(\ref{eq_claim_1}).
\end{proof}

 %%% Uncomment this line and comment out the ``thebibliography'' section below to use the external .bib file (using bibtex) .





% \end{thebibliography}

\end{document}